\documentclass[lettersize,journal]{IEEEtran}
\usepackage{amsmath,amsfonts}
\usepackage{algorithm}
\usepackage{array}
\usepackage{textcomp}
\usepackage{stfloats}
\usepackage{url}
\usepackage{verbatim}
\usepackage{graphicx}
\usepackage{cite}
\usepackage{algorithmicx}
\usepackage{algpseudocode}
\usepackage{multirow}
\usepackage{booktabs}
\usepackage{colortbl} 
\usepackage{xcolor}
\usepackage{float}
\usepackage{subfig}
\usepackage{amssymb}
\usepackage{bbding}
\usepackage{array}

\def \b {{\mathbf{b}}}
  
  \def \e {\mathbf{e}}

\def \h {\mathbf{h}} \def \H {\mathbf{H}} 
  
  \def \0 {\mathbf{0}}
\def \p {\mathbf{p}} 
\def \o {\mathbf{o}}

  \def \O {\mathbf{O}}

\def \x {\mathbf{x}}

\def \Z {\mathbf{Z}} \def \z {\mathbf{z}}
 \def \Eps \mathbf{\varepsilon}

\def \ca {{\mathcal A}}

  \def \cn {{\mathcal N}}
\def \cl {{\mathcal L}}  
  
 \def \cs {{\mathcal S}} 
 \def \cc {{\mathcal C}} \def \cg {{\mathcal G}}

\makeatletter

\newcommand{\Rmnum}[1]{\expandafter\@slowromancap\romannumeral #1@}
\makeatother

\newcommand{\btheta}{\boldsymbol{\theta}}

\newcommand{\bomega}{\mbox{\boldmath $\omega$}}

\def \method {CoMe}
\newcommand{\ie}{{i}.{e}., }
\newcommand{\eg}{{e}.{g}., }

\begin{document}

\title{Towards Long-Tailed Recognition for Graph Classification via Collaborative Experts}

\author{Si-Yu Yi$^*$, Zhengyang Mao$^*$, Wei Ju$^\dag$, Yong-Dao Zhou, Luchen Liu, Xiao Luo, and Ming Zhang$^\dag$% <-this % stops a space
\thanks{This paper is partially supported by the National Natural Science Foundation of China with Grant (NSFC Grant Numbers 62306014, 62106008, 62276002 and 12131001) as well as the China Postdoctoral Science Foundation with Grant No. 2023M730057.}
\IEEEcompsocitemizethanks{\IEEEcompsocthanksitem Si-Yu Yi and Yong-Dao Zhou are with School of Statistics and Data Science, Nankai University, Tianjin, 300071, China.
(e-mail: siyuyi@mail.nankai.edu.cn, ydzhou@nankai.edu.cn)
\IEEEcompsocthanksitem Zhengyang Mao, Wei Ju, Luchen Liu and Ming Zhang are with National Key Laboratory for Multimedia Information Processing, School of Computer Science, Peking University, Beijing, 100871, China.
(e-mail: mao.zhengyang.cn@gmail.com, juwei@pku.edu.cn, liuluchen@pku.edu.cn, mzhang$\_$cs@pku.edu.cn)
\IEEEcompsocthanksitem Xiao Luo is with Department of Computer Science, University of California, Los Angeles, 90095, USA. (e-mail: xiaoluo@cs.ucla.edu)
\IEEEcompsocthanksitem Corresponding authors: Wei Ju and Ming Zhang 
\IEEEcompsocthanksitem Si-Yu Yi and Zhengyang Mao contribute equally to this paper.  %with order determined by flipping a coin.
}
}



\maketitle

\begin{abstract}
Graph classification, aiming at learning the graph-level representations for effective class assignments, has received outstanding achievements,
% along with the development of graph neural networks (GNNs). The success 
which heavily relies on high-quality datasets that have balanced class distribution. 
% However, most real-world graph data naturally presents a long-tailed form, where the head classes occupy much more samples than the tail classes, while the graph-level classification over long-tailed data still remains largely unexplored. 
In fact, most real-world graph data naturally presents a long-tailed form, where the head classes occupy much more samples than the tail classes, it thus is essential to study the graph-level classification over long-tailed data while still remaining largely unexplored.
However, most existing long-tailed learning methods in visions fail to jointly optimize the representation learning and classifier training, as well as neglect the mining of the hard-to-classify classes. 
Directly applying existing methods to graphs may lead to sub-optimal performance, since the model trained on graphs would be more sensitive to the long-tailed distribution due to the complex topological characteristics.
% \revA{However, existing long-tailed learning methods fail to jointly optimize the representation learning and classifier training in the long-tailed problem, both of which are sensitive to the imbalanced class distribution, especially for the graph-structured data with complex topological characteristics. Along with the neglect of the hard-to-classify classes, it would lead to sub-optimal performance if the existing methods are directly applied to the graphs.}
% Moreover, existing long-tailed learning methods mostly concentrate on the class-rebalancing strategies while failing to jointly optimize the representation learning and classifier training in the long-tailed problem, and neglect the attention to the hard-to-classify classes, which leads to sub-optimal performance. 
% Besides, the graph-level classification over long-tailed data still remains largely unexplored. 
Hence, in this paper, we propose a novel long-tailed graph-level classification framework via \underline{\textbf{Co}}llaborative \underline{\textbf{M}}ulti-\underline{\textbf{e}}xpert Learning (CoMe) to tackle the problem. 
% It equilibrates the contributions of head and tail classes in balanced contrastive learning along with individual-expert classifier training based on hard class mining. 
To equilibrate the contributions of head and tail classes, we first develop balanced contrastive learning from the view of representation learning, and then design an individual-expert classifier training based on hard class mining.
In addition, we execute gated fusion and disentangled knowledge distillation among the multiple experts to promote the collaboration in a multi-expert framework. Comprehensive experiments are performed on seven widely-used benchmark datasets to demonstrate the superiority of our method CoMe over state-of-the-art baselines.
\end{abstract}


\section{Introduction}
\IEEEPARstart{G}{raph-structured} data \cite{kipf2016semi,xie2022active,ju2023comprehensive} is ubiquitous in a variety of domains, such as social networks, protein-protein interaction networks, and citation networks. Graph classification \cite{xie2022semisupervised,li2022semi,ju2023tgnn,luo2023towards}, as one of the most fundamental tasks in data mining for graphs, has attracted significant attention. It attempts to predict the class label and the property of each graph in a dataset. Promising applications include property prediction for quantum mechanics \cite{hao2020asgn} and functional assessment of chemical compounds \cite{kojima2020kgcn}.

Graph neural networks (GNNs) \cite{kipf2016semi,hamilton2017inductive,xie2022active,zhang2021hyperbolic} have become one of the most prominent approaches in graph representation learning and achieved remarkable progress for graph classification. The key idea of GNNs is to iteratively propagate and aggregate the information from the neighbors of each node in a graph for generating node-level representations \cite{gilmer2017neural}. Afterward, a readout function \cite{lee2019self,ying2018hierarchical} integrates all of the node representations into a graph-level representation, which is then fed into a classifier to predict the graph label. Hence, the learned graph-level representation can incorporate the characteristics of the topological structure and reveal the whole semantic information of the graph, which works well for the downstream graph classification task. In spite of this, the huge success of GNNs is typically built on high-quality datasets of having a roughly balanced distribution. Nevertheless, real-world datasets commonly exhibit long-tailed class distributions, where a large portion of tail classes occupy a limited number of data whereas few head classes have most of the data. For example, in the Cora citation network, the proportion of the instances in head class \emph{Neural Network} is about $26.8\%$, while those in the tail classes \emph{Rule Learning} and \emph{Reinforcement Learning} are only about $7.9\%$ and $4.8\%$ respectively. In such scenarios, directly adopting GNNs on long-tailed graph datasets may lead to notorious prediction bias and significant performance degradation, since the model is prone to being dominated by the head class while attention to tail classes is easily overlooked. Consequently, the high-frequency head classes may achieve impressive predictive performance, while the low-resource tail classes receive unsatisfactory accuracy, thereby hindering the strength of GNNs learned from long-tailed graph data.

To tackle the problem caused by long-tailed data distribution, existing studies have proposed lots of methods \cite{chawla2002smote, he2009learning, ren2018learning,  wang2020long, cui2021parametric,maurya2017distributed} in computer vision, which fall into three categories: re-sampling, re-weighting, and ensemble learning. The re-sampling strategy \cite{chawla2002smote, guo2021long} aims to balance the data distribution in the head and tail classes, including over- and under-sampling. Over-sampling replicates existing samples in the tail classes, whereas under-sampling discards some samples from the head classes to reduce the imbalance. 
However, re-sampling can lead to over-fitting or under-fitting problems, since improper re-sampling may excessively utilize the minority samples or abandon useful information in the majority samples. 
The re-weighting strategy \cite{cao2019learning, tan2020equalization} modifies the weights of the losses from different classes to make the long-tailed data contribute properly to training. 
But plain re-weighting strategy could benefit classifier learning while hurting representation learning, because it may under-represent the head classes and cause unstable training \cite{kang2019decoupling,alshammari2022long}.
%both of which are essentially important in the graph classification task.
Different from the formers, ensemble learning \cite{wang2020long, li2022trustworthy} combines multiple expert networks from a complementary perspective to obtain reliable and robust predictions, whose works have achieved satisfactory progress. 
Nevertheless, most current ensemble learning methods lack mutual supervision among different experts and knowledge transfer is also deficient.
% To facilitate long-tailed learning for graph-structured data, many GNN-based methods have been proposed for node classification. \cite{liu2021tail} hinged on the novel concept of transferable neighborhood translation to model the variable ties between a target node and its neighbors. \cite{qu2021imgagn} presented a generative adversarial graph network to simulate both the attributive and topological distributions such that the number of nodes in different classes can be balanced. 
% \cite{hu2022graphdive} learned multi-view graph representations and combined multi-view experts to investigate the relationship between topological structure and class imbalance. 

Despite the wealth of the researches in visions, however, to the best of our knowledge, the long-tailed graph-level classification is yet rarely explored. Thus, it is essential to pay insight into the task for meeting the demand of the practical applications. 
When dealing with long-tailed graphs, because of the complex topology structure and the ubiquitous over-smoothing issue caused by the message passing in GNN-based encoder \cite{park2021graphens}, the classification model trained on graphs would be more sensitive to the long-tailed class distribution and it thus is necessary to improve all the components of the model that can be affected by the long-tailed distribution, such as the representation learning and classifier training. Nevertheless, due to the shortcomings of the existing methods in visions discussed above, directly applying them to the long-tailed graph data cannot solve such problems roundly and effectively, which may result in sub-optimal performance. Moreover, many existing methods focus on the training of the hard samples by re-sampling or re-weighting while ensemble learning lays emphasis on aggregating different perspectives by multiple experts for more comprehensive data exploration. These strategies overlook the selection of the hard-to-classify classes, which leads to many samples being indistinguishable from the non-target classes (i.e., hard classes) but still having high confidence. Instead, the selection of the hard classes for all samples is actually a more refined and comprehensive way to distinguish samples that are difficult to classify accurately and acquire higher classification accuracy. Hence, it is highly desirable to proposed a novel method to jointly optimize representation learning and classifier learning as well as effectively capture hard classes for long-tailed graphs.

% Despite the promising achievements, the majority of existing long-tailed learning approaches still suffer from three key limitations: 
% (i) \textbf{Inability to balance the representation learning and classifier learning in long-tailed problem.} Many methods pay more attention to classifier learning while ignoring the influence of long-tailed data on representation learning, both of which are crucial for long-tailed classification. In some scenarios, the boost of the classifier may hurt the performance of representation learning \cite{kang2019decoupling}. 
% (ii) \textbf{Lack of ability to capture hard classes.} Existing algorithms often focus on selecting hard samples for training, this however cannot explicitly capture hard classes for all samples, which results in many samples being indistinguishable from the non-target classes (i.e., hard classes), while still having high confidence. 
% (iii) \textbf{Rare exploration on the long-tailed graph classification task.} To the best of our knowledge, previous research has mainly concentrated on node classification under the long-tailed setting, while graph-level classification is yet rarely explored. Therefore, it is highly desirable to jointly optimize representation learning and classifier learning as well as effectively capture hard classes. Furthermore, it is essential to pay insight into the long-tailed graph-level classification for meeting the demand of the practical applications, where most of the data fall into the long-tailed distribution. 

To address these issues, based on \underline{\textbf{Co}}llaborative \underline{\textbf{M}}ulti-\underline{\textbf{e}}xpert Learning, we propose a novel framework named \method{} for long-tailed graph classification. The key idea of \method{} is to propose tailored balanced contrastive learning along with individual-expert classifier training to jointly optimize the representation learning and classifier learning, % better balancing the contributions of the head and tail classes, 
and then fuse and distill the multiple expert networks from both global and local views for stronger collaboration capability. Specifically, \method{} first introduces tailored balanced contrastive learning to alleviate the class imbalance for representation learning, which can effectively balance the contributions of the head and tail classes on the contrastive loss. Then, we propose the balanced predicted probability in the classifier learning for each expert from both global and local perspectives %based on the hard class mining 
to alleviate the influence of the sample sizes in different classes on embedded representations
 and enhance the hard class mining. 
%From the global view, the contributions of head and tail classes in the classifier training can be balanced, whereas, from the local view, hard class learning is enhanced. 
%The integration of the two techniques can effectively prevent the expert model from being overwhelmed by the head classes. 
Moreover, to fully benefit from multi-expert/ensemble learning, we fuse different expert models by gating functions to increase the diversity of the whole training network and meanwhile, we perform knowledge distillation among experts in a disentangled manner to encourage them learn extra knowledge from others and decouple the effects of the predictions for the target class and non-target classes. 
Comprehensive experiments are conducted to show that the proposed method can greatly improve the performance of the long-tailed graph classification compared with existing state-of-the-art methods over multiple benchmark datasets. Moreover, the combination of tailored representation learning and classifier training  
%our tailored multi-expert framework and the strategy of the hard-to-classify class mining 
is highly effective for dealing with imbalanced settings. 
To summarize, the main contributions of our work are as follows:

\begin{itemize}
\item This paper studies long-tailed graph-level classification under the multi-expert learning framework, which jointly optimizes the representation learning and classifier training, and explores the organic fusion and effective cooperation among expert networks.

\item We propose a framework to balance the contributions of the head and tail classes on both representation learning and classifier learning. We explicitly capture the hard-to-classify classes for all samples and equip the multi-expert learning with gated fusion and disentangled knowledge distillation to enhance the long-tailed learning.

\item Extensive experiments are performed on various commonly used datasets to validate the superiority of the proposed approach against existing state-of-the-art models.
\end{itemize}

% The rest of this paper is organized as follows. 
% Section~\ref{sec::related} reviews the related work and Section~\ref{sec::definition} introduces some preliminaries. In Section~\ref{sec::model}, we describe the details of our proposed method. Experiments on several benchmark datasets are conducted in Section~\ref{sec::experiment}. 
% Finally, Section~\ref{sec::conclusion} concludes the paper and visions for future work. 

\section{Related Work}
\label{sec::related}

In this section, we briefly review the related works in three aspects, namely graph-level classification, long-tailed learning, and contrastive learning.

\subsection{Graph-level Classification}
% \noindent\textbf{Graph-Level Classification}
Graph-level classification is one of the most critical problems in the graph domain, which aims at predicting the class label of the entire graph. Existing algorithms for graph classification can be broadly categorized into graph kernel-based methods and GNN-based methods. The core of classic graph kernels is to decompose each graph into substructures (e.g., graphlets \cite{shervashidze2009efficient}, subtrees \cite{shervashidze2011weisfeiler}, or shortest paths \cite{kashima2003marginalized}) to measure the similarity between two graphs. Recent works have focused on designing expressive GNNs \cite{kipf2016semi,hamilton2017inductive,zhang2021hyperbolic,chen2022fedge} and achieved remarkable success. 
The key idea of GNNs \cite{kipf2016semi,hamilton2017inductive,xie2022active,ju2022ghnn,ju2022kgnn} is to iteratively update the node feature according to its neighbor nodes with pooling methods \cite{lee2019self,ying2018hierarchical} to integrate all node representations and characterize meaningful representation of the whole graph. Our paper goes further and explores a challenging and under-explored scenario, i.e., long-tailed graph-level classification.

\subsection{Long-tailed Learning}
Long-tailed learning aims to alleviate the impact of class imbalance on model training, and there are currently three main strategies to address this practical problem: re-sampling~\cite{chawla2002smote, he2009learning, buda2018systematic, zhao2021graphsmote, qu2021imgagn, yi2023model}, re-weighting~\cite{cao2019learning, cui2019class, huang2019deep, liu2021tail, zhang2021distribution}, and ensemble learning~\cite{wang2020long, xiang2020learning, cai2021ace, li2022trustworthy, hu2022graphdive}.
For the re-sampling group, SMOTE~\cite{chawla2002smote} aims to rebalance the data distribution by generating new samples and performing interpolation in the tail classes, which belongs to an over-sampling approach. GraphSMOTE~\cite{zhao2021graphsmote} extends SMOTE to the graph domain by encoding and synthesizing new samples based on the similarity between nodes, and training an edge generator to model relationship information. For the re-weighting group, LDAM~\cite{cao2019learning} proposes a label-distribution-aware margin loss inspired by minimizing a margin-based generalization bound. DisAlign~\cite{huang2019deep} develops an adaptive calibration function that enables the adjustment of classification scores for individual data points. As for ensemble learning, RIDE~\cite{wang2020long} designs an effective strategy to reduce model variance and bias as well as mitigate computational costs via dynamic expert routing. NCL and NCL++~\cite{li2022nested, tan2023ncl++}  propose two complementary modules, which emphasize individual supervised learning for each expert and knowledge transfer among experts respectively, within a nested framework for comprehensive representation learning. Our work inherits the advantages of ensemble learning and is dedicated to organic cooperation and supervision among experts to promote effective long-tailed learning.
% LFME~\cite{xiang2020learning} aggregates knowledge from multiple experts and proposes a knowledge distillation framework to learn a unified student model.

\subsection{Contrastive Learning}
% \smallskip\noindent\textbf{Contrastive Learning}
Contrastive learning (CL) learns the common and discriminative attributes by a contrasting principle among the positive and negative pairs. Many approaches have been proposed with competitive performance~\cite{chen2020simple, he2020momentum, chen2020improved, khosla2020supervised,  ren2021label}. SimCLR~\cite{chen2020simple} combines the data augmentations and a learnable nonlinear transformation in CL to learn the representations. MoCo~\cite{he2020momentum} builds a dynamic dictionary with a queue and a moving-averaged encoder to facilitate contrastive self-supervised learning. SupCon~\cite{khosla2020supervised} extends the contrastive approach to the fully-supervised setting, which allows for effective leverage of the label information. MoCov2~\cite{chen2020improved} improves MoCo by using a multi-layer perceptron (MLP) projection head and more data augmentations to ease the burden on the batch size in training. SACC~\cite{deng2023strongly} incorporates strong and weak augmentations into instance- and cluster-level CL for deep clustering. Moreover, there are many recent methods extending CL to graph domains~\cite{you2020graph, zhong2023contrastive,ju2023unsupervised,luo2022clear,luo2022dualgraph,ju2023glcc,yuan2023learning}. GraphCL~\cite{you2020graph} designs four types of graph augmentations to incorporate various priors for effective CL. CGCN~\cite{zhong2023contrastive} proposes a semi-supervised contrastive loss to maximize the homogeneity of the original topology graph and the self-adaptive one. For long-tailed graph classification, our paper focuses on designing tailored supervised contrastive learning that balances the head and tail classes to learn effective graph-level representations. 
% that integrates the label information and the function of balancing head and tail classes to learn the graph-level representations. multi-layer perceptron (MLP)

% \subsection{Knowledge Distillation}

\section{Preliminaries}\label{sec::definition}

In this section, we first briefly present the basic notations and formal terminologies for the long-tailed graph dataset. Then, we provide the problem definition for the long-tailed graph-level classification task and give an introduction to the GNN-based classifier. 

\smallskip
{\bf Notations.} 
% Given a graph dataset with $M$ classes denoted by $\cg=\{G_i=(V_i, E_i, \X_i), y_i\}_{i=1}^N$, where $G_i$ is the $i$-th graph, $V_i$ is the node set, $E_i$ is the edge set, $\X_i$ is the attribute matrix and $y_i\in\{1,\ldots, M\}$ is the class label of $G_i$. 
Given a graph dataset $\cg=\{G_i, y_i\}_{i=1}^N$ with $M$ classes, where $G_i$ is the $i$-th graph and $y_i\in\{1,\ldots, M\}$ is the ground-truth class label of $G_i$. 
Let $N_j$ denote the number of graphs in the $j$-th class and assume that $N_1\geq N_2 \geq \cdots \geq N_M$ % w.l.o.g.
without loss of generality. 
The imbalance factor (IF) of the dataset is defined as $N_1/N_M$ to measure the extent of class imbalance. 
%Following Zipf's law \cite{newman2005power}, a dataset is long-tailed if $N_i=N_1 \times i^{-\mu}$ for $i=1,\ldots, K$, where $\mu$ is a parameter controlling the degree of data imbalance measured by IF. For real-world data, the long-tailed class distribution probably does not exactly, but approximatively follow Zipf's law. 
Let ${\rm p}(G | y)$ be the probability density function of graph $G$ conditioned on the class $y$. 
A dataset is long-tailed if the following relationships hold: 
\begin{equation}\label{LT}
\begin{cases}
   \int {\rm p}(G|y=j') {\rm d}G \geq \int {\rm p}(G|y=j'') {\rm d}G, & \forall ~ j' \leq j'', \\
    \lim_{j \rightarrow \infty} \int {\rm p}(G|y=j){\rm d}G = 0, &
\end{cases}
\end{equation}
where $j', j'' \in\{1,\ldots, M\}$  are indices of class labels. Eq.~\eqref{LT} indicates that for any given class index $j' \leq j''$, the number of samples in the $j'$-th class is larger than the number of samples in the $j''$-th class. In other words, the above formula reflects that the class size successively decays with the class index increasing and the probability finally approaches zero in the last few classes. Under the long-tailed setting in Eq.~\eqref{LT}, the classes can be divided into head, medium, and tail classes based on different numbers of graphs. The number of graphs in head classes is far more than that of tail classes. 

\smallskip
% {\bf Problem Definition.} For the long-tailed graph-level classification problem, the target is to train an unbiased classifier based on the long-tailed dataset $\cg$. The classifier needs to learn effective and discriminative representations for tail classes in a low-resource and imbalanced setting, such that it is not overwhelmed by the abundant head-class graphs and is able to classify both head and tail classes correctly with respect to certain balanced metrics. For example, the mean accuracy is an appropriate choice to emphasize both the head and tail classes, i.e., $\text{mean\_acc} = \frac{1}{M} \sum_{j=1}^M \text{acc}_j$, where $\text{acc}_j$ is the accuracy calculated over the graphs in the $j$-th class. Further, the learned classifier should have a powerful generalization ability on a balanced test dataset. 
{\bf Problem Definition.} The target of long-tailed graph-level classification is to train an unbiased classification model based on the long-tailed graph dataset. The model needs to learn effective and discriminative representations for all the graphs, such that it is not overwhelmed by the abundant head classes and is able to correctly classify the graphs in all classes. Further, the trained classifier should have a powerful generalization ability on a balanced test dataset.

\smallskip
{\bf GNN-based Classifier.} To obtain effective probability assignments on the classes for each graph, GNN is the most popular strategy in graph representation learning. We begin with leveraging GNN as the encoder to acquire the whole graph representation, which reasonably captures the node attribute and structure information. Specifically, the propagation rule in a layer of GNN is 
\begin{equation*}%\label{GNN}
\h_v^{(l+1)} = \cc_{\btheta}^{(l)}\left(\h_v^{(l)}, \ca_{\btheta}^{(l)}(\{\h_{v'}^{(l)}\}_{v'\in \cn(v)})\right),
\end{equation*}
where $\h_v^{(l)}$ is the learned representation of node $v$ in Graph $G$ at the $l$-th layer, $\cn(v)$ is the neighbors of node $v$, $\cc_{\btheta}^{(l)}$ and $\ca_{\btheta}^{(l)}$ are the combination and aggregation functions at the $l$-th layer, respectively. Based on the node-level representations obtained by a $L$-layer GNN, a pooling operation $READOUT$ is performed to merge them into a graph-level representation of graph $G$, which is formulated as 
\begin{equation}\label{Grep}
\h = READOUT(\{\h_{v'}^{(L)}: v' \in V\}), 
\end{equation}
where $V$ is the node set of graph $G$. Then, the graph representation of fusing attributive and topological information can be used for graph-level classification, where the representation is fed into a classifier, such as a multi-layer perceptron (MLP) followed by a softmax function, to derive the probability assignment of graph $G$. 

However, direct employment of the GNN-based classifier on a long-tailed dataset easily results in prediction bias, since the model may be dominated by the high-frequency head classes. 
% %with a majority of data. 
Hence, ensuring the high accuracy of the GNN-based classifier for both head and tail classes is what we focus on.

\section{The proposed method}\label{sec::model}

\begin{figure*}
    \centering \includegraphics[width=1\textwidth]{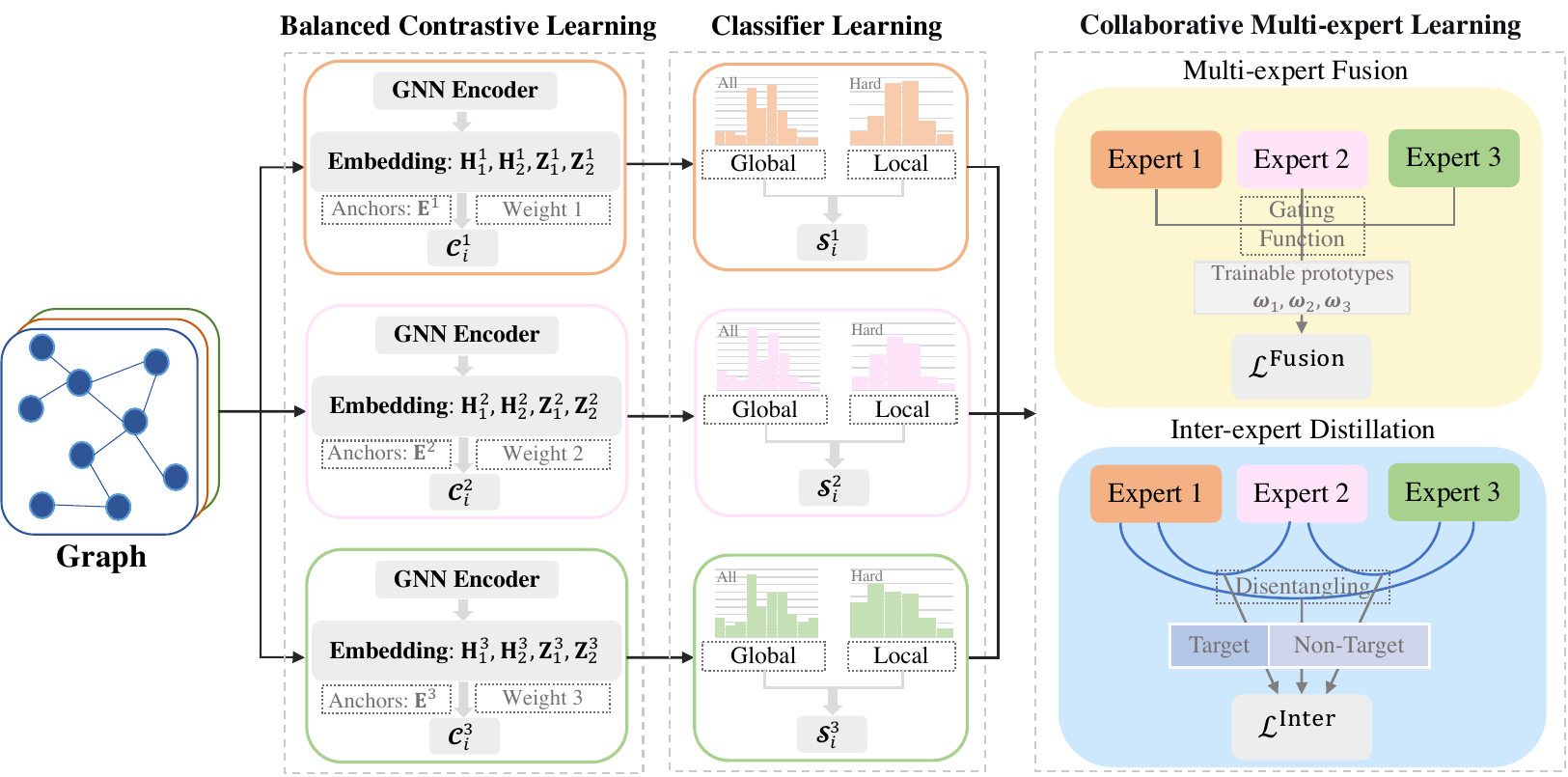}
    \caption{Framework overview of the proposed \method{}. Balanced contrastive learning and classifier learning from both global and local views are performed to jointly guide the individual-expert training. Then the experts are integrated by gated multi-expert fusion and disentangled inter-expert knowledge distillation for long-tailed graph classification.} 
    \label{fig:framework}
\end{figure*}

In this section, we introduce our proposed framework named \method{}, which explores the challenging long-tailed graph classification in the context of multi-expert learning. Our \method{} mainly contains four modules, i.e., balanced contrastive learning of individual expert, individual-expert classifier learning, multi-expert fusion module, and inter-expert distillation module. Figure \ref{fig:framework} presents the framework overview of the proposed method. In the following, we show the four parts of our framework \method{} in detail. 

\subsection{Balanced Contrastive Learning of Individual Expert}

The high performance of collaborative multi-expert learning is based on the fact that each expert network has excellent learning ability. Thus, we first discuss the representation learning of the individual expert. To accommodate the long-tailed dataset, we propose a tailored balanced contrastive learning (BCL) to generate discriminative representations for each expert, which can facilitate a balance between the contributions of head and tail classes and alleviate the problem caused by insufficient samples in the tail classes.

Contrastive learning \cite{chen2020simple} is a self-supervised learning that extracts meaningful representations by maximizing the similarity of positive pairs and minimizing the correlation of negative pairs based on two augmented view of the data. To achieve this, data augmentation plays a vital role in contrastive learning to enrich the dataset and acquire the positive/negative pairs. To increase the diversity of the experts, we leverage different graph augmentation strategies in different expert networks. Specifically, we consider four strategies following \cite{you2020graph}: 
\begin{itemize}

%\item {\bf Attribute Masking.} For each value in the attribute matrix of each graph, we randomly choose a few nodes and mask their corresponding attributes. This strategy slightly disturbs the node features, which would not essentially change the semantic information. 

\item {\bf Attribute Masking.} We randomly mask a few entries of the node attributes in the graph. This strategy slightly disturbs the node features, which would not essentially change the semantic information. 

% \item {\bf Node Dropping.} We delete a certain portion of nodes in each graph, where the dropping probability follows a uniform distribution. For each dropped node, the connected edges without affecting the semantics are also removed. 
\item {\bf Node Dropping.} We randomly delete a certain portion of nodes in each graph along with the connected edges removed, the probability of a node being removed follows a uniform distribution.

\item {\bf Edge Perturbation.} We randomly add or delete several edges from each graph. It is premised on the assumption that the semantic information is resistant to variations in edge connection patterns.

\item {\bf Subgraph.} We utilize the random walk algorithm to sample a subgraph from the original graph with the assumption that the main semantics can be preserved in the local structure to some extent. 

\end{itemize}

Assume that a total of $K$ experts are employed and in the $k$-th expert model, we obtain two augmented datasets $\{\cg_1^k, \cg_2^k\}$ through different augmentations on the original $N$ graphs, where the subscripts (\ie $a=1,2$) denote the $a$-th augmented view. The graphs in $\cg_1^k$ and $\cg_2^k$ are fed into the shared GNN encoder to learn the graph-level representations as Eq. \eqref{Grep}, which are denoted by $\H_1^k=\{\h_1^{1,k},\ldots, \h_N^{1,k}\}$ and $\H_2^k=\{\h_1^{2,k},\ldots, \h_N^{2,k}\}$, respectively. As the traditional contrastive learning \cite{chen2020simple}, the learned representations are then mapped into a shared space by an MLP to compute the contrastive loss, where the mapped embeddings are denoted by $\Z_1^k=\{\z_1^{1,k},\ldots, \z_N^{1,k}\}$ and $\Z_2^k=\{\z_1^{2,k},\ldots, \z_N^{2,k}\}$, respectively. To balance the contributions of the head and tail classes in contrastive learning without sacrificing the accuracies of the head classes, we introduce the trainable anchors $E(k)=\{\e_1^k, \ldots, \e_M^k\}$ as learnable class representations. Specifically, we define the balanced contrastive loss for the $i$-th graph and the $k$-th expert as 

\begin{small}
\begin{equation}\label{loss_bcl_i_k}
\cc^{k}_{i} = - \sum_{\z_{+}\in P(i) \cup \e_{y_i}^k} w(\z_{+}) \log \frac{\exp(\z_{+} \cdot T(\h_i^{1,k})/\tau)}{\sum_{\z_j \in A(i) \cup E(k)} \exp(\z_j \cdot T(\h_i^{1,k})/\tau)}, \\
\end{equation}
\end{small}
with
\begin{align*}%\label{AP}
A(i)  & = \{\z_j \in \Z_1^k \cup \Z_2^k\} \backslash \z_i^{1,k}, \\
P(i) & = \{\z_j \in A(i): y_j = y_i\}, 
\end{align*}
\begin{align*}
\z \cdot T(\h_i^{1,k}) & =
\begin{cases}
     \z \cdot \z_i^{1,k}, & \z \in A(i), \\
     \z \cdot  \h_i^{1,k}, & \z \in E(k),
\end{cases}\\
% \end{align*}
% \begin{equation}\label{loss_bcl_i_k}
% \begin{gathered}
% \mathcal{C}_i^k=-\sum_{\mathbf{z}_{+} \in P(i) \cup \mathbf{e}_{y_i}} w\left(\mathbf{z}_{+}\right) \log \frac{\exp \left(\mathbf{z}_{+} \cdot T\left(\mathbf{h}_i^1\right) / \tau\right)}{\sum_{\mathbf{z}_j \in A(i) \cup \mathbf{E}^k} \exp \left(\mathbf{z}_j \cdot T\left(\mathbf{h}_i^1\right) / \tau\right)}, ~ \text { with } \\
% A(i)=\left\{\mathbf{z}_j \in \mathbf{Z}_1^k \cup \mathbf{Z}_2^k\right\} \backslash \mathbf{z}_i^1,~ P(i)=\left\{\mathbf{z}_j \in A(i): y_j=y_i\right\},~ \mathbf{z} \cdot T\left(\mathbf{h}_i^1\right)= \begin{cases}\mathbf{z} \cdot \mathbf{z}_i^1, & \mathbf{z} \in A(i), \\
% \mathbf{z} \cdot \mathbf{h}_i^1, & \mathbf{z} \in \mathbf{E}^k\end{cases}
% \end{gathered}
% \end{equation}
% where $\h_i^1\in \H_1^k$, $\z_i^1\in \Z_1^k$, 
%are the embedded representations for the $i$-th graph in $\cg_1^k$, 
% \begin{equation*}%\label{BCL}
w(\z_+) & =
\begin{cases}
     \alpha, & \z_+ \in P(i), \\
     1, & \z_+ = \e_{y_i}^k,
\end{cases}
\end{align*}
% \end{equation*}
where $\tau$ is the temperature hyper-parameter, $w(\z_+)$ is the weight function, $y_i$ is the true label of the $i$-th graph, and $\alpha \in [0,1]$ is a weight parameter. The set $A(i)$ contains all the graph representations from the $k$-th expert except the representation $\z_i^{1,k}$ for the $i$-th graph from the first augmented view, and the set $P(i)$ includes the representations of the graphs that belong to class $y_i$ in $A(i)$.
%Then the total balanced contrastive loss (BCL) for the $k$-th expert is written as 
%\begin{equation}\label{loss_bcl_k}
%\cc^k=\sum_{i=1}^N \cc^{k}_{i}.
%\end{equation}

Next, we theoretically illustrate the usefulness of the defined loss $\cc^k_i$ for representation learning on long-tailed data. 
Denote $W_{i}$ as the cardinality of $P(i)$, 
$P(i)=\{\z_1^+,\z_2^+,\ldots, \z_{W_{i}}^+\}$, $p_q^+ = {\exp(\z_{q}^+ \cdot T(\h_i^{1,k})/\tau)}/{\sum_{\z_j \in A(i) \cup E(k)} \exp(\z_j \cdot T(\h_i^{1,k})/\tau)}$ for $q=1,2,\ldots,W_{i}$, $p_{\e}^+={\exp(\e_{y_i}^k \cdot T(\h_i^{1,k})/\tau)}/{\sum_{\z_j \in A(i) \cup E(k)} \exp(\z_j \cdot T(\h_i^{1,k})/\tau)}$, and $p^+ = p_1^++\cdots+p_{W_{i}}^+ +p_{\e}^+ \in [0,1]$. 
Then the balanced contrastive loss in Eq. \eqref{loss_bcl_i_k} can be rewritten as 
\begin{equation*}
\cc_i^k = - \alpha (\log p_1^++\cdots+\log p_{W_{i}}^+) - \log p_{\e}^+. 
\end{equation*}
To achieve the minimum of $\cc_i^k$, according to the Lagrange multiplier method \cite{bertsekas2014constrained}, the Lagrange equation is formed as 
\begin{align*}
F  = & - \alpha (\log p_1^++\cdots+\log p_{W_{i}}^+) - \log p_{\e}^+ \\
& + \lambda ( p_1^+ +\cdots+p_{W_{i}}^+ +p_{\e}^+ - p^+),
\end{align*}
where $\lambda$ is the Lagrange multiplier. We take the first-order derivatives of $F$ with respect to $\lambda, p_q^+, p_{\e}^+$ and set the derivatives equal to 0, which are formulated as 
\begin{align*}%\label{derive}
\frac{\partial F}{\partial \lambda} & =  p_1^+ +\cdots+p_{W_{i}}^+ +p_{\e}^+ - p^+ = 0, \nonumber\\
\frac{\partial F}{\partial p_q^+} &  =  -\alpha \frac{1}{p_q^+} + \lambda = 0, \\
\frac{\partial F}{\partial p_{\e}^+} & =  - \frac{1}{p_{\e}^+} + \lambda = 0. \nonumber
\end{align*}
By jointly solving the above equations, we obtain that $p_q^+=\alpha p^+ / (\alpha W_{i}+1)$ and $p_{\e}^+ = p^+ / (\alpha W_{i}+1)$. Hence, if $p^+ = 1$, $\cc_i^k$ achieves its minimum with 
\begin{align*}% \label{PaCo}
\frac{\exp({\z_{+}}\cdot T(\h_i^{1,k})/\tau)}{\sum_{\z_j \in A(i) \cup E(k)} \exp(\z_j \cdot T(\h_i^{1,k})/\tau)}  = & \frac{1}{W_{i} + 1/\alpha}   %\text{ for any } \z \in P(i), 
\end{align*}
satisfied for any $\z_{+} \in P(i)$, which is the probability that two graphs in the same class are a positive pair. 

As for the original supervised contrastive loss (SupCon) in \cite{khosla2020supervised}, which is defined as 
\begin{equation*}%\label{SupCon_loss}
\bar{\cc}^{k}_{i} = - \sum_{\z_{+}\in P(i)} \log \frac{\exp(\z_{+} \cdot T(\h_i^{1,k})/\tau)}{\sum_{\z_j \in A(i)} \exp(\z_j \cdot T(\h_i^{1,k})/\tau)}. 
\end{equation*}
It can be similarly obtained that when $\bar{\cc}^{k}_{i}$ reaches its minimum, for any $\z_{+} \in P(i)$, we have
\begin{equation}\label{SupCon}
  \frac{\exp({\z_{+}}\cdot T(\h_i^{1,k})/\tau)}{\sum_{\z_j \in A(i)} \exp(\z_j \cdot T(\h_i^{1,k})/\tau)}  =  \frac{1}{W_{i}}.   %\text{ for any } \z \in P(i).
\end{equation}
We can observe in Eq. \eqref{SupCon} that a larger $W_{i}$ yields a smaller probability that two graphs in the same class are a positive pair, which causes a greater contribution from the negative log of Eq. \eqref{SupCon} to $\bar{\cc}^{k}_{i}$. 
It implies that when the training in SupCon converges, the high-frequency head classes have a greater impact on the loss than the low-resource tail classes since the samples in head classes possess larger $W_{i}$ ($=2N_{y_i}-1$), which inevitably inhibits the performance of the tail classes. 

Let $W_{h}$ and $W_{t}$ be the cardinalities of $P(i_h)$ and $P(i_t)$, where $i_h$ and $i_t$ are the head- and tail-class graphs, respectively. 
By introducing the class anchors $E(k)$ and the weight $w(\z_+)$ in Eq. \eqref{loss_bcl_i_k}, our BCL reduces the difference of the probabilities that two graphs are a positive pair in the head and tail classes from $1/W_t-1/W_h$ to $1/(W_t+1/\alpha)-1/(W_h+1/\alpha)$. When $\alpha$ is smaller, the difference could be smaller, which balances the contributions of the head and tail classes to the loss in Eq. \eqref{loss_bcl_i_k}. 
In addition, with the decrease of $\alpha$, the weight of the contrast between the graph and the corresponding class representation in Eq. \eqref{loss_bcl_i_k} would be higher, which improves the contrast intensity between them, thereby bearing the ability to push graphs in the same class close to each other and implicitly benefits hard class learning. Hence, the proposed balanced contrastive loss enhances the performance of the learned representation over long-tailed graph data.

\subsection{Individual-expert Classifier Learning}

Based on the learned embedding $\H^k$ from the original graphs in the aforementioned balanced contrastive learning, we feed it into a classifier network such as an MLP, to obtain the logit vectors $\O^{k}=\{\o_1^{k}, \ldots, \o_N^{k}\}$. Then we use the softmax-like function to derive the predicted probability assignment for each graph sample. If the naive softmax function is adopted, for the $i$-th graph and the $k$-th expert, the predicted probability of assigning to the $j$-th class is defined as 
\begin{equation}\label{softmax}
    \bar{p}_{i,j}^k = \frac{\exp(o_{i,j}^k)}{\sum_{m=1}^M \exp(o_{i,m}^k)}, 
\end{equation}
where $o_{i,j}^k$ is the $j$-th entry in the logit vector $\o_i^{k}$. Apparently, the naive softmax function cannot mitigate the impact of the long-tailed distribution on the classifier, which inevitably leads to the learned classification model dominated by the high-frequency head classes. It implies that the learned logits from the classifier implicitly incorporate the effects of the sample sizes for different classes, which is not beneficial to the testing graph whose ground-truth label belongs to the tail class. 
Hence, we need to eliminate the effects of the sample sizes for different classes on the logits, which can indirectly equilibrate their impacts on the trained classifier. 

To this end, we leverage the idea of Bayesian inference and incorporate class frequency into the predicted probability by treating it as the prior information. Specifically, we propose the balanced predicted probability as follows, 
\begin{align}\label{BPPA}
p_{i, j}^k= & {\rm p}(y=j|\x=\o_i^k)=\frac{{\rm p}(\x=\o_i^k|y=j){\rm p}(y=j)}{{\rm p}(\x=\o_i^k)} \nonumber\\
=& \frac{N_j \exp(o_{i,j}^k)}{\sum_{m=1}^M N_m \exp(o_{i,m}^k)},
\end{align}
where $y$ is the label of $\x$ and $N_j$ is the number of graphs in the $j$-th class of the dataset. The probability ${\rm p}(\x=\o_i^k|y=j)$ is defined as $\bar{p}_{i,j}^k$ in Eq. \eqref{softmax} 
% calculated by performing softmax function on $\o_i^k$, \ie ${\rm p}(\o_i^k|y=j)={\exp(o_{i,j}^k)}/{\sum_{m=1}^M \exp(o_{i,m}^k)}$ 
and ${\rm p}(y=j)=N_j/N$. 
When the graph dataset is completely balanced, \ie $N_1=\cdots=N_M$, the defined $p_{i,j}^k$ naturally degenerates into the vanilla softmax function. 
Intuitively, the incorporation of the prior information ${\rm p}(y=j)$ in ${p}_{i, j}^k$ decouples the effects of sample size $N_{j}$ and the logit $\o_i^k$ on the prediction result. 
% which effectively alleviates the influence of the sample sizes in different classes on the embedded representations. 
Moreover, in the testing stage, a common principle is to conduct softmax operation on the learned logit vector for the testing sample to acquire the predicted probability. Thus, accompanied by a supervised loss (\eg cross entropy) based upon the true labels and the balanced probabilities defined in Eq. \eqref{BPPA}, the training process can be well supervised 
% from a global perspective 
to alleviate the influence of the sample sizes for different classes on the logits, which effectively avoids the prediction performance being overwhelmed by sufficient head-class samples.

\smallskip
\textbf{Hard-to-classify Class Mining.} Further, under the long-tailed setting, it is ubiquitous that the predicted class for the sample in the tail class is not the ground-truth class but with a high predicted score~\cite{leevy2018survey}. Hence, motivated by \cite{li2022nested, tan2023ncl++}, additional attention shall be paid to hard class mining (HCM). 
%to improve the prediction accuracy. 
%for the tail-class samples. 
% Here we treat all graphs equally, and 
For the $i$-th graph and the $k$-th expert, we explicitly extract the hard classes $\Omega_i^k$ by selecting classes with the top-$M_{\rm hard}$ largest logits of the $i$-th row in $\O^k$ and the ground-truth class, which is formulated as
\begin{equation}\label{TopHC}
\Omega_i^k = {\rm TopHC}\{j: j\neq y_i \} \cup \{y_i\}.
\end{equation}
Then, we define the balanced predicted probability focusing on the hard classes by integrating $\Omega_i^k$ into Eq. \eqref{BPPA}, which derives 
\begin{equation}\label{HBPPA}
\tilde{p}_{i, j}^k = \frac{N_j \exp(o_{i,j}^k)}{\sum_{m\in \Omega_i^k} N_m \exp(o_{i,m}^k)} \text{ for any } j \in \Omega_i^k. 
\end{equation}
Eq. \eqref{HBPPA} presents a fine-grained characterization of the probability assignment among the hard classes. % By providing a local view, 
It enables more targeted and effective supervised classifier learning, which ultimately leads to improved prediction accuracy for tail classes. 
% It provides a local view for subsequent supervised classifier learning, which is beneficial to improve the prediction accuracy of tail classes. 

% \subsection{Intra-expert Learning Module}\label{traM}

Based on the above discussion, to derive the expert network with strong discrimination ability for graph classification, we conduct classifier learning for each expert from both global and local perspectives. From the global view, we resort to ${p}_{i,j}$ in Eq. \eqref{BPPA} to empower the ability of balancing the contributions of head and tail classes in training, whereas from the local view, we enhance the hard class learning with the help of $\tilde{p}_{i, j}^k$ in Eq. \eqref{HBPPA}. Formally, for the $i$-th graph and the $k$-th expert, the supervised loss of individual classifier is defined as
\begin{equation*}%\label{traL_i}
\cs_i^k = - (\log({p}_{i,y_i}^k) + \log(\tilde{p}_{i,y_i}^k) ). 
\end{equation*}

\subsection{Multi-expert Fusion Module}\label{traM}

In addition to individual learning for each expert, how to organically combine the losses from different experts is also the focus of multi-expert learning  %Individual learning leverages the divide-and-conquer strategy to assign different experts to learn different knowledge. 
since the graphs in different classes may have heterogeneous interactions with different expert networks. Here we employ gating functions to control the fusion process, which can be regarded as learnable weights and trained along with expert networks. By the law of total probability, the joint prediction mechanism with multiple experts is formulated as 
\begin{equation*}%\label{con}
{\rm p}(y=y_i | \O) = \sum_{k=1}^K {\rm p}(e_i=k| \o_i^k){\rm p}(y=y_i|e_i=k, \o_i^k),
\end{equation*}
where $\o_i^k= (o_{i,1}^k,\ldots, o_{i,M}^k)^{\top}$, $\O$ is formed by stacking the logit vectors of $K$ experts, $e_i$ is a latent variable indicating the expert index for the $i$-th graph, ${\rm p}(e_i=k| \o_i^k)$ is the gating function of the $k$-th expert with $\sum_{k=1}^K {\rm p}(e_i=k| \o_i^k) =1$ satisfied, and ${\rm p}(y=y_i|e_i=k, \o_i^k)$ represents the predicted probability of the $k$-th expert. For the gating function, we parameterize it as an input-dependent soft assignment based on the cosine similarity between the embedded logit $\o_i^k$ and the trainable gating prototype $\bomega^k$, which has the form as
\begin{equation*}%\label{gate}
{\rm p}(e_i=k| \o_i^k) = \frac{\exp(\o_i^k\cdot \bomega^k / \kappa)}{\sum_{k=1}^K \exp(\o_i^k\cdot \bomega^k / \kappa)}, 
\end{equation*}
where $\kappa$ is the temperature hyper-parameter. The value ${\rm p}(y=y_i|e_i=k, \o_i^k)$ can be substituted by the balanced predicted probability ${p}_{i,y_i}^k$ or its variant $\tilde{p}_{i,y_i}^k$. Hence, with the adoption of both ${p}_{i,y_i}^k$ and $\tilde{p}_{i,y_i}^k$ from global and local views, the fused supervised loss (FSL) of classifiers can be equipped as 
\begin{equation*}
\cl^{\rm FSL} = \sum_{i=1}^N \sum_{k=1}^K {\rm p}(e_i=k| \o_i^k) \cs_i^k. 
\end{equation*}
%where ${\rm p}(e_i=k| \o_i^k)$ is defined in Eq. \eqref{gate}. 

In addition, following an analogous pipeline, we can similarly fuse the balanced contrastive loss $\cc_i^k$ in Eq. \eqref{loss_bcl_i_k} under different experts, which derives the fused contrastive loss (FCL) as 
\begin{equation*}%\label{loss_bcl}
\cl^{\rm FCL} = \sum_{i=1}^N \sum_{k=1}^K {\rm p}(e_i=k| \o_i^k) \cc_i^k. 
\end{equation*}
Thoroughly, for the $N$ graphs and $K$ experts, the total fusion loss for multiple experts is defined as
\begin{align}
\label{traL}
\cl^{\rm Fusion} &= \cl^{\rm FSL} + \eta \cl^{\rm FCL} \nonumber \\  
& = \sum_{i=1}^N \sum_{k=1}^K {\rm p}(e_i=k| \o_i^k) (\cs_i^k+\eta \cc_i^k), 
\end{align}
where $\eta$ is the pre-defined hyper-parameter to adjust the weight between $\cs_i^k$ and $\cc_i^k$. 
With the adaptive fusion of multiple experts,  the diversity of the whole training network is increased, which can jointly boost the performance of the graph-level classification task on long-tailed data. 

\subsection{Inter-expert Distillation Module}

Besides the fusion of the experts, the knowledge distillation among the experts is also significant to allow each expert network to learn extra signals from others and achieve information sharing. We employ the Kullback-Leibler (KL) divergence-like metric to support the inter-expert distillation. 

Here we define the balanced predicted probability among the non-target classes as
$$\breve{p}_{i, j}^k = \frac{N_j \exp(o_{i,j}^k)}{\sum_{m\neq y_i} N_m \exp(o_{i,m}^k)}, j\in \{1,\ldots,M\}\backslash \{y_i\}.$$
We easily have $\breve{p}_{i, j}^k = {p}_{i, j}^k/(1-{p}_{i,y_i}^k)$. In the standard KL divergence, the distance between the probability assignments of two experts for the $i$-th graph can be formulated as  
\begin{align}\label{KL}
\text{KL}({\rm \p}_i^{k} || {\rm \p}_i^{q}) = & {p}_{i,y_i}^k \log\left(\frac{{p}_{i,y_i}^k}{{p}_{i,y_i}^q} \right) +  \sum_{j\neq y_i} {p}_{i, j}^k \log\left(\frac{{p}_{i, j}^k}{{p}_{i,j}^q} \right) \nonumber\\
= & {p}_{i,y_i}^k \log\left(\frac{{p}_{i,y_i}^k}{{p}_{i,y_i}^q} \right) + (1-{p}_{i,y_i}^k)\log\left(\frac{1-{p}_{i,y_i}^k}{1-{p}_{i,y_i}^q} \right) \nonumber \\
 & + (1-{p}_{i,y_i}^k) \sum_{j\neq y_i} \breve{p}_{i, j}^k \log\left(\frac{\breve{p}_{i,y_i}^k}{\breve{p}_{i,y_i}^q} \right) \nonumber\\
\triangleq & \text{KL}({\rm \b}_i^{k} || {\rm \b}_i^{q}) + (1-{p}_{i,y_i}^k)\text{KL}(\breve{\rm \p}_i^{k} || \breve{\rm \p}_i^{q}),
\end{align}
where ${\rm \p}_i^{k}=({p}_{i,1}^{k}, \ldots, {p}_{i,M}^{k})^{\top}$, ${\rm \b}_i^{k}=({p}_{i,y_i}^k, 1-{p}_{i,y_i}^k)^{\top}$ is the binary probability assignment vector of whether it belongs to the target class (i.e., the ground-truth class), and $\breve{\rm \p}_i^{k} = (\breve{p}_{i,1}^k, \ldots,\breve{p}_{i,y_i-1}^k, \breve{p}_{i,y_i+1}^k,\ldots,\breve{p}_{i,M}^k)^{\top}$ represents the probability assignment among the non-target classes. $\text{KL}({\rm \b}_i^{k} || {\rm \b}_i^{q})$ measures the similarity of the binary probability assignments of the target class in different expert networks, while $\text{KL}(\breve{\rm \p}_i^{k} || \breve{\rm \p}_i^{q})$ represents the similarity of the probability assignments among the non-target classes in different expert networks. The weight $1-{p}_{i,y_i}^k$ is negatively correlated with the prediction performance. Under the long-tailed setting, we can find from Eq. \eqref{KL} that due to the natural weight, the high prediction confidence on the graphs in the head classes would inevitably suppress the information sharing among the non-target classes between two experts; whereas, the low prediction confidence on the samples in the tail classes would spontaneously hinder the mutual learning of target knowledge between the experts, which further inhibits the prediction performance of the tail classes under the framework of collaborative multi-expert learning. Hence, to better leverage the advantage of multi-expert learning, we propose a disentangled metric (DKL) for knowledge distillation, which is defined as
\begin{equation*}%\label{DKL}
\text{DKL}({\rm \p}_i^{k} || {\rm \p}_i^{q}; \beta_1, \beta_2) = \beta_1 \text{KL}({\rm \b}_i^{k} || {\rm \b}_i^{q}) + \beta_2\text{KL}(\breve{\rm \p}_i^{k} || \breve{\rm \p}_i^{q}), 
\end{equation*}
where $\beta_1, \beta_2$ are weight hyper-parameters. With the flexible weight, DKL effectively avoids mutual inhibition between the distillations of the target and non-targets to promote high-performance cooperation.

Moreover, we analogously analyze the KL divergence $\text{KL}(\tilde{\p}^k_i || \tilde{\p}^q_i)$ between two different experts' probability assignments that focus on the hard classes with $\tilde{\p}^k_i=(\tilde{p}^k_{i,1},\ldots,\tilde{p}^k_{i,M})^{\top}$ and generate a disentangled version denoted by $\text{DKL}(\tilde{\p}_i^{k} || \tilde{\p}_i^{q}; \beta_1,\beta_2)$. Based on $\text{DKL}({\rm \p}_i^{k} || {\rm \p}_i^{q}; \beta_1,\beta_2)$ and $\text{DKL}(\tilde{\p}_i^{k} || \tilde{\p}_i^{q}; \beta_1,\beta_2)$, we incorporate both the global and local perspectives as Section \ref{traM} to further promote the distillation performance for the long-tailed data. Accordingly, the total inter-expert distillation loss is defined as
\begin{align}\label{terL}
\cl^{\rm Inter} = & \sum_{k=1}^K \sum_{q\neq k}^K \sum_{i=1}^N ( \text{DKL}({\rm \p}_i^{k} || {\rm \p}_i^{q}; \beta_1,\beta_2)   \nonumber \\
 &  + \text{DKL}(\tilde{\rm \p}_i^{k} || \tilde{\rm \p}_i^{q}; \beta_1,\beta_2) ). 
\end{align}

\subsection{Joint Optimization Module for Graph Classification}\label{class}

% To obtain excellent performance for such data on graph classification tasks, we propose pertinent schemes in terms of contrastive learning and supervised learning, which leverage information from data and labels, respectively. 
Graph classification is essentially a supervised task and we focus on the long-tailed data in this paper. Toward this end, we incorporate the multi-expert learning framework to increase the diversity of the whole network, where individual learning, multi-expert fusion and inter-expert distillation are discussed to jointly promote the performance of the long-tailed graph classification task. 

Formally, we unite the multi-expert fusion loss $\cl^{\rm Fusion}$ in Eq. \eqref{traL} and the inter-expert distillation loss $\cl^{\rm Inter}$ in Eq. \eqref{terL} to jointly optimize our proposed framework \method{}, where the total loss $\cl$ is 
\begin{equation}\label{totalL}
\cl = \cl^{\rm Fusion} + \epsilon \cl^{\rm Inter}, 
\end{equation}
where $\epsilon$ is the pre-defined hyper-parameter to adjust the influence of $\cl^{\rm Fusion}$ and $\cl^{\rm Inter}$ on training. 

After converges, we feed each test instance into the network to obtain the logit vectors $\{\o_{\rm test}^k\}_{k=1}^K$ and output the predicted probability vector ${\p}_{\rm test}$ by performing the softmax operation on the averaged logit $\bar{\o}_{\rm test} = \sum_{k=1}^K \o_{\rm test}^k/K$. The whole process of our proposed \method{} is summarized in Algorithm \ref{code1}.

\begin{algorithm}[t]
	\caption{The pseudo-code of the proposed \method{}}
	\label{code1}
	\begin{algorithmic}[1]
	\Require Graph dataset $\cg=\{G_i,y_i\}_{i=1}^N$; Class number $M$; 	Expert number $K$; Maximum iterations $T_{max}$; 
	\Ensure Classification result $y$; 
	\State Initialize the parameters in balanced contrastive learning; 
	\For{$t=1$ to $T_{max}$}
	\State Obtain $\{\cg_1^k, \cg_2^k\}_{k=1}^K$ by graph augmentation; 
	\State Update the embeddings $\{\H_1^k, \H_2^k, \Z_1^k, \Z_2^k\}_{k=1}^K$ and the logits $\{\O^k\}_{k=1}^K$ by the GNN-based encoder; 
	\State Extract the hard classes $\Omega_i^k$ in Eq. \eqref{TopHC} traversing all the graphs and experts; 
	\State Calculate the losses $\cl^{\rm Fusion}$ and $\cl^{\rm Inter}$ in Eq. \eqref{traL} and Eq. \eqref{terL},  respectively;
	\State Conduct backpropagation and update the whole network in \method{} by minimizing $\cl$ in Eq. \eqref{totalL};
	\EndFor
	\State Obtain the predicted probability $\p_{\rm test}$ with the averaged logit $\bar{\o}_{\rm test}$ on the test instance; 
	\State \Return $y$;
	\end{algorithmic}
\end{algorithm}

\subsection{Computational Complexity Analysis}

%Assume that GraphSAGE \cite{hamilton2017inductive} is adopted as the GNN backbone encoder and the batch size is $B$. Given a dataset $\cg=\{G_i=(V_i, E_i, \X_i)\}_{i=1}^N$, the time complexity of GraphSAGE is $O(\sum_{j=1}^B |V_j|\prod_{i=1}^D S_i)$, where $S_i$ is the neighborhood sample size in different layer and $D$ is the search depth. 
%For the scalability of large-scale datasets, 
In training, we adopt the mini-batch stochastic gradient descent to optimize our method. Assume that the batch size is $B$ and the complexity of producing embeddings and logits from the GNN-based encoder is $O(W)$. For $K$ experts, we calculate the balanced contrastive loss in $O(Kd(\sum_{i=1}^B |A(i)| + BM))$ time, where $A(i)$ is defined in Eq. \eqref{loss_bcl_i_k}, $d$ is the dimension of the embeddings and $M$ is the number of classes. The complexity of classifier learning is $O(KBM)$. Moreover, the time complexities of the multi-expert fusion and inter-expert distillation are $O(K(M+B))$ and $O(KB(M-1))$, respectively. Hence, the total computational complexity of our approach is $O(KW + Kd\sum_{i=1}^B |A(i)| + KBMd)$.

\section{Experiments}\label{sec::experiment}

In this section, we first introduce the experimental settings which include benchmark datasets, compared baselines, and implementation details of the proposed method. 
Then we conduct experiments to validate the effectiveness of \method{}. We aim to answer the following research questions. 

\begin{itemize}
    \item {\bf RQ1:} Does our proposed \method{} outperform baseline methods in long-tailed graph classification?
    \item {\bf RQ2:} How do different components of \method{} contribute to the overall classification performance?
    \item {\bf RQ3:} How do the hyperparameters in \method{} affect the final classification performance?
    \item {\bf RQ4:} How does disentangled knowledge distillation affect the classification performance of \method{}?
\end{itemize}

\subsection{Experimental Setup}

% \subsubsection{Benchmark Datasets}
\noindent\textbf{Datasets.}
We evaluate the proposed \method{} against several baselines on seven publicly accessible datasets from various fields, including (a) social networks: COLLAB \cite{yanardag2015deep}, (b) synthetic: Synthie \cite{morris2016faster}, (c) bioinformatics: ENZYMES \cite{schomburg2004brenda}, and (d) computer vision: MNIST \cite{dwivedi2020benchmarking}, Letter-high \cite{riesen2008iam}, Letter-low \cite{riesen2008iam}, and COIL-DEL \cite{riesen2008iam}. Among the seven datasets utilized in the experiments, COLLAB and ENZYMES datasets are natural graph datasets derived from real-world data, while the Synthie dataset is a synthetic graph dataset. Additionally, the visual-world MNIST dataset is represented as graphs by setting the superpixels as nodes and their spatial
relations as edges, 
% partitioning each image into superpixels, 
while the Letter-high, Letter-low, and COIL-DEL datasets are transformed into graph structures by representing the lines in letters as undirected edges and considering the endpoints of these lines as nodes, all originating from real-world image datasets. To ensure that the datasets follow Zipf's law exactly  \cite{reed2001pareto}, we processed the original training sets into standard long-tailed datasets with different imbalance factors (IFs), while the validation and test sets remained to be balanced. We choose distinct IFs for each dataset to ensure the number of training samples in the tail class with the least samples falls within the range of $2$ to $4$.
% the least frequent class

% \subsubsection{Compared Baselines}
\smallskip
\noindent\textbf{Baselines.}
To demonstrate the efficacy of our proposed framework, we compare our \method{} with a range of competitive long-tailed learning baselines. Below we give a brief introduction to nine baseline models, which can be divided into four main categories, \ie, data re-balancing, loss re-weighting, information augmentation, and contrastive learning based methods. Among the baseline methods considered, Graph augmentation, G$^2$GNN, and GraphCL are specifically designed for long-tailed learning on graphs, whereas CB loss, LACE loss, SupCon, and SBCL are adapted from approaches originally developed for long-tailed classification tasks in visual world.

\begin{itemize}
    \item GraphSAGE \cite{hamilton2017inductive} serves as a basic GNN encoder in our implementation, where we utilize the mean aggregator to aggregate feature information. This method is used as the base encoder for \method{} and other baselines.
    \item Over-sampling \cite{chawla2003c4} technique often incorporates repeating samples from tail classes randomly as a means of making datasets balanced.
    \item CB loss \cite{cui2019class} is a loss re-weighting approach at the class level. To tackle the training problem for imbalanced data, CB loss adds class-balanced weighting to the loss function for class $i$ inversely, which is proportional to the effective number of samples.
    \item LACE loss \cite{menon2020long} is another re-weighting technique that adjusts the prediction probabilities using the label frequencies, the LACE loss function employs adjustment to the logits during the model inference phase.
    \item Graph augmentation \cite{yu2022graph} is a popular technique in graph representation learning that enhances model generalization and generates extra training data. We use over-sampling to increase training data and select either edge permutation or node dropping as one of the two fundamental topological augmentations.
    \item G$^2$GNN \cite{wang2022gog} measures the graph kernel-based similarity between different graph samples to construct a Graph-of-Graph (GoG), which links graphs with their k-nearest neighbors. After constructing the kNN graph, neighboring graph representations are aggregated together via the GoG propagation on the established kNN graph.
    \item GraphCL \cite{you2020graph} proposes four distinct strategies to augment input graphs and learn graph-level representations, which aims to maximize the mutual information between the original graph and its augmented variants.
    \item SupCon \cite{khosla2020supervised} is an extended version of the contrastive loss, which leverages label information effectively. SupCon allows more than one view to be positive so that views of the same label can be attracted to each other in the embedding space.
    \item SBCL \cite{hou2023subclass} tackles the limitations of contrastive learning by clustering each head class into multiple subclasses of comparable sizes to the tail classes, thereby achieving subclass balance and learning more balanced representation space for long-tailed data.
\end{itemize}
% These baseline methods can be broadly categorized into four perspectives: (a) Data re-sampling methods: over-sampling \cite{chawla2003c4}; (b) Loss re-balancing methods: class-balanced (CB) loss \cite{cui2019class} and logit adjusted cross-entropy (LACE) loss \cite{menon2020long}; (c) Information augmentation methods: graph augmentation \cite{yu2022graph} and G$^2$GNN \cite{wang2022gog}; (d) Contrastive learning based methods: graph contrastive learning (GraphCL) \cite{you2020graph} and supervised contrastive learning (SupCon) \cite{khosla2020supervised}.

% \subsubsection{Implementation Details}
\smallskip
\noindent\textbf{Implementation Details.}
In all experiments, we used GraphSAGE \cite{hamilton2017inductive} as the GNN backbone encoder, with a two-layer MLP classifier. To optimize all models, we utilized the Adam optimizer with a fixed learning rate of $0.0001$ and a batch size of $32$. For our proposed \method{}, we set the expert number $K$ to $3$ to balance performance and efficiency. We also set the temperature hyper-parameter $\tau$ and the contrast weight $\alpha$ for the BCL module to $0.2$ and $0.05$, respectively. For joint training, we set the contrast weight $\eta$ to $1.0$ and the inter-expert distillation weight $\epsilon$ to $0.6$. Moreover, we tuned the hyper-parameters of HCM number $M_{\rm hard}$, and the DKL hyper-parameters $\beta_{1}$ and $\beta_{2}$ for each dataset. 
% For the baseline methods, we set the re-weighting hyper-parameter to $0.99$ for the CB loss and the scaling temperature to $1.0$ for the LACE loss.
All baseline models are implemented in PyTorch using the open-sourced code provided by the original paper. 
%We 
%adopt the mean accuracy over the classes as the evaluation metric and 
The top-$1$ average accuracy of $10$ run times is employed for evaluation.

\begin{table*}[t]
\begin{center}
\caption{Overall performance ($\%$) with various IFs on seven benchmark datasets for long-tailed graph classification. The best results are shown in boldface and the second-best results are underlined.}
\label{table:overall}
\tabcolsep=7pt
\resizebox{0.98\textwidth}{!}{ %
{\small
\begin{tabular}{lcccccccccccccc}
\toprule
\multirow{2}{*}{Model}  &\multicolumn{2}{c}{COLLAB} &\multicolumn{2}{c}{Synthie} &\multicolumn{2}{c}{ENZYMES}  &\multicolumn{2}{c}{MNIST} &\multicolumn{2}{c}{Letter-high} &\multicolumn{2}{c}{Letter-low} & \multicolumn{2}{c}{COIL-DEL}\\
& IF=10 & IF=20 & IF=15 & IF=30 & IF=15 & IF=30 & IF=50 & IF=100 & IF=25 & IF=50  & IF=25 & IF=50 & IF=10 & IF=20\\ 
\midrule
GraphSAGE &63.07 &53.33 &34.74  &30.25  &30.66  &25.16 &68.67 &63.46 &51.06  &42.16  &86.00  &84.32  &38.80  &31.32 \\
Over-sampling &72.33 &70.25 &35.25  &33.50  &32.33  &28.50  &64.69 &59.78 &53.62  &44.20  &88.48  &86.72  &39.20  &26.96 \\
\midrule
CB loss &68.78 &65.85 &34.75  &30.75  &32.19  &26.83 &68.85 &63.40 &53.76  &45.06  &87.46  &85.44  &41.72  &32.34 \\
LACE loss &68.33 &64.77 &33.25  &30.85  &31.16  &25.50 &69.72 &64.59 &47.46  &38.94  &87.89  &84.69  &41.96  &32.18 \\
\midrule
Augmentation &72.85 &71.14 &39.37  &35.37  &32.08  &26.75 &72.18 &68.17 &49.28  &42.36  &88.32  &86.40  &38.18  &30.80 \\
G$^2$GNN$_{n}$ &73.94 &71.89 &38.08  &27.94  &35.00  &29.17 &70.91 &66.73  &\underline{58.91}  &51.12  &89.49  &87.98  &38.32  &27.98 \\
G$^2$GNN$_{e}$  &\underline{74.50} &\underline{72.76}   &40.19  &37.53  &35.83  &29.50 &73.69 &70.31 &58.85  &49.96  &89.84 &87.80  &39.18  &31.06 \\ 
\midrule
GraphCL &69.33 &67.36 &40.25   &36.25   &36.66  &29.83 &69.37 &65.12 &57.34  &48.93  &89.28  &87.89  &42.02   &33.19 \\
SupCon  &69.25 &67.14 &40.34   &37.25   &37.08  &30.67 &69.76 &64.88 &57.29  &48.93  &89.12  &87.36  &42.93 &34.20 \\
SBCL &71.63 &69.12 &\underline{42.08}  &\underline{38.19}   &\underline{37.63}  &\underline{32.41}&\underline{75.06} &\underline{72.12} &57.73  &\underline{51.38}  &\underline{90.54}  &\underline{89.03} &\underline{44.78}  &\underline{37.64} \\
\midrule
\method{} &\textbf{76.88} &\textbf{74.40} &\textbf{42.25}  &\textbf{38.50}  &\textbf{38.00}  &\textbf{33.50}  &\textbf{76.24} &\textbf{72.80} &\textbf{63.42}  &\textbf{54.12}  &\textbf{91.67}  &\textbf{90.56}  &\textbf{46.56}  &\textbf{38.88} \\
% \midrule
% Improve $\uparrow$  &+3.19\%  &+2.25\%  &+12.17\%  &+10.57\%  &+2.48\%  &+9.22\%  &+7.66\%  &+5.87\%  &+2.03\%  &+2.93\% &+13.11\%  &+13.68\%\\
\bottomrule
\end{tabular}
}
}
\end{center}
\end{table*}

\subsection{Overall Comparison (RQ1)}
In this section, we evaluate the performance of \method{} on long-tailed graph classification and compare it with various baseline methods. Table \ref{table:overall} presents the experimental results on seven benchmarks with different IFs. Based on the quantitative results, the following observations can be made:
\begin{itemize}
    \item %After a comprehensive analysis of the classification accuracy on all seven datasets, 
    Based on the classification accuracies on all seven datasets, 
    we observed a sharp decrease in the performance of all methods as the long-tailedness between head and tail classes increases. This indicates that GNNs are highly susceptible to long-tailed distribution and degrade severely in such long-tailed settings.
    \item For all four types of baseline methods, information augmentation and contrastive learning approaches mainly focus on learning better representations, while loss re-balancing methods aim to balance the classifier during training. From the results, it can be observed that existing baselines fail to balance the representation learning and classifier training in long-tailed learning, which leads to sub-optimal classification performance. Overall, information augmentation approaches outperform re-balancing approaches on most datasets as they incorporate additional knowledge to enrich the tail classes. Moreover, contrastive learning baselines demonstrate relatively stable performance across all datasets.
    \item %Based on the results presented in the table, 
    %it can be concluded that our proposed \method{} outperforms all baseline approaches and achieves the best performance on all seven datasets. 
    From the table, it can be concluded that our \method{} achieves the best performance on all seven datasets compared with all the baselines. 
    It is mainly attributed to the fact that for effectively addressing the long-tailed problem in graph data, our \method{} not only enhances representation learning with balanced contrastive learning but also promotes classifier balancing with balanced predicted probability. Compared to methods that directly adapt from image long-tailed classification approaches, which mainly concentrate on rebalancing the classifier (CB loss, LACE loss) or improving representations (SupCon, SBCL), our approach excels in effectively addressing the specific long-tailed problem encountered in graph data. 
     In addition, \method{} outperforms other competitors with a large margin under strict imbalance settings (COIL-DEL with IF=10 and 20), where most classes have fewer than 5 training instances. This highlights the importance of employing hard class mining to derive the expert network with strong discrimination ability. 
     % \revA{Moreover, our method effectively addresses the long-tailed problem in graph data by simultaneously balancing the data distribution at the representation and classifier levels. Compared to methods that directly adapt from image long-tailed classification approaches, which mainly concentrate on rebalancing the classifier (CB loss, LACE loss) or improving representations (SupCon, SBCL), our approach excels in effectively addressing the specific long-tailed problem encountered in graph data.}
\end{itemize}

\begin{table}[t]
\begin{center}
\caption{Comparison with several variants for ablation study ($\%$) on Letter-high and ENZYMES datasets (Multi-expert framework: MeF; Balanced contrastive learning: BCL; Hard class mining: HCM; Gated Multi-expert fusion: GMeF; Disentangled Inter-expert distillation: DIeD).}
\label{table:ablation}
\tabcolsep=1.5pt
%\resizebox{0.6\columnwidth}{!}{
{\small
\begin{tabular}{c|ccccc|cccc}
\toprule
 &\multicolumn{1}{c}{\multirow{2}{*}{MeF}}     &\multicolumn{1}{c}{\multirow{2}{*}{BCL}}     &\multicolumn{1}{c}{\multirow{2}{*}{HCM}}    &\multicolumn{1}{c}{\multirow{2}{*}{GMeF}}    &\multicolumn{1}{c|}{\multirow{2}{*}{DIeD}}  &\multicolumn{2}{c}{Letter-high}  &\multicolumn{2}{c}{ENZYMES} \\
 &\multicolumn{1}{c}{} &\multicolumn{1}{c}{} &\multicolumn{1}{c}{} &\multicolumn{1}{c}{} &\multicolumn{1}{c|}{} &Accuracy &$\Delta$ &Accuracy &$\Delta$ \\
\midrule
% $M_1$ &\Checkmark & & & & &54.92 &- &35.00 &- \\
% $M_2$ &\Checkmark &\Checkmark & & & &62.34 &+7.42 &36.93 &+1.93 \\
% $M_3$ &\Checkmark &\Checkmark &\Checkmark & & &62.54 &+7.62 &37.00 &+2.00 \\
% $M_4$ &\Checkmark &\Checkmark &\Checkmark &\Checkmark & &62.85 &+7.93 &37.13 &+2.13 \\
% $M_5$ &\Checkmark &\Checkmark &\Checkmark &\Checkmark &\Checkmark &63.42 &+8.50 &38.00 &+3.00 \\
$M_1$ &\Checkmark & & & & &54.92 &- &35.00 &- \\
$M_2$ &\Checkmark & &\Checkmark & & &57.47 &+2.55 &35.67 &+0.67 \\
$M_3$ &\Checkmark &\Checkmark & & & &62.34 &+7.42 &36.93 &+1.93 \\
$M_4$ &\Checkmark &\Checkmark &\Checkmark & & &62.54 &+7.62 &37.00 &+2.00 \\
$M_5$ &\Checkmark &\Checkmark &\Checkmark &\Checkmark & &62.85 &+7.93 &37.13 &+2.13 \\
$M_6$ &\Checkmark &\Checkmark &\Checkmark & &\Checkmark &62.65 &+7.73 &37.33 &+2.33 \\
$M_7$ &\Checkmark &\Checkmark &\Checkmark &\Checkmark &\Checkmark &63.42 &+8.50 &38.00 &+3.00 \\
\bottomrule
\end{tabular}
}
\end{center}
\end{table}

\subsection{Ablation Study (RQ2)}

% \subsubsection{Ablation study on major components}
\noindent\textbf{Ablation study on major components.}
We conduct an ablation analysis for the key components in our \method{}, \ie multi-expert framework (MeF), balanced contrastive learning (BCL), hard class mining (HCM), gated multi-expert fusion (GMeF), and disentangled inter-expert distillation (DIeD). We analyze the effect of each component by adding them to the base model until obtaining the complete method. In all cases, the classification results in the training process are generated by the balanced predicted probability. Table \ref{table:ablation} demonstrates the results of the ablation study. The base model ($M_1$) includes only the model ensemble, without any of the other key components.
% Adding BCL ($M_2$) results in a significant improvement in accuracy on both datasets, indicating the importance of balanced contrastive learning for better representation learning. The addition of HCM ($M_3$) and gating prototypes ($M_4$) further improves the accuracy, demonstrating the effectiveness of hard class mining and gating prototypes in enhancing the discrimination ability of the experts and dynamically fusing multiple experts. Finally, the addition of disentangled inter-expert distillation ($M_5$) also provides an improvement in accuracy, indicating that the use of knowledge distillation between expert models in a disentangled manner can effectively boost knowledge sharing between experts.
Adding HCM ($M_2$) and BCL ($M_3$) to the baseline model significantly improves accuracy on both datasets. Particularly, $M_3$ shows a substantial improvement, emphasizing the importance of balanced contrastive learning for effective representation learning. Furthermore, the combination of HCM and BCL ($M_4$) leads to further accuracy improvements. The introduction of gating prototypes ($M_5$) enhances discrimination ability and enables the dynamic fusion of multiple experts, resulting in higher accuracy. Incorporating disentangled inter-expert distillation ($M_6$) also contributes to accuracy improvement by facilitating effective knowledge sharing between experts through disentangled knowledge distillation. Finally, when all key components are combined ($M_7$), the model achieves the best performance, underscoring the collective impact of the key components on overall accuracy.

% \subsubsection{Ablation study on balanced contrastive learning and balanced predicted probability}
\smallskip
\noindent\textbf{Ablation study on BCL and BPP.}
To analyze the effect of balanced contrastive learning, we carry out an experiment by substituting BCL with unsupervised contrastive learning (UCL) and supervised contrastive learning (SCL). Table \ref{table:bcl} shows the experimental results on two datasets. From the table, we can observe that the classification accuracy of using both BCL and BPP is higher than combining UCL or SCL with BPP, indicating that BCL learns better representation in the long-tailed settings. However, employing BCL without BPP results in a minor drop in classification accuracy. These results suggest that the combination of BCL and BPP can lead to significant improvements in both representation learning and classifier learning, which can ultimately enhance the overall performance of the model.

\begin{table}[t]
\begin{center}
\caption{Effectiveness analysis ($\%$) of balanced contrastive learning and balanced predicted probability on the Synthie and Letter-high datasets (Unsupervised contrastive learning: UCL; Supervised contrastive learning: SCL; Balanced contrastive learning: BCL; Balanced predicted probability: BPP).}
\label{table:bcl}
\tabcolsep=4pt
{\small
% \resizebox{0.96\columnwidth}{!}{
\begin{tabular}{cccc|cccc}
\toprule
 \multicolumn{1}{c}{\multirow{2}{*}{UCL}}     &\multicolumn{1}{c}{\multirow{2}{*}{SCL}}     &\multicolumn{1}{c}{\multirow{2}{*}{BCL}}    &\multicolumn{1}{c|}{\multirow{2}{*}{BPP}}  &\multicolumn{2}{c}{Synthie}  &\multicolumn{2}{c}{Letter-high} \\
 \multicolumn{1}{c}{} &\multicolumn{1}{c}{} &\multicolumn{1}{c}{} &\multicolumn{1}{c|}{} &Accuracy &$\Delta$ &Accuracy &$\Delta$ \\
\midrule
 & & & &36.25 &- &53.60 &- \\
\Checkmark & & &\Checkmark &37.35 &+1.10 &60.37 &+6.77 \\
 &\Checkmark & &\Checkmark &40.23 &+3.98 &61.65 &+8.05 \\
 & &\Checkmark & &39.62 &+3.37 &60.69 &+7.09 \\
 & &\Checkmark &\Checkmark &42.25 &+6.00 &63.42 &+9.82 \\
\bottomrule
\end{tabular}}
\end{center}
\end{table}

\begin{figure*}[t]
\centering
	\subfloat[COIL-DEL]{
    \label{subfig:param1}
    \includegraphics[width = 0.24\textwidth]{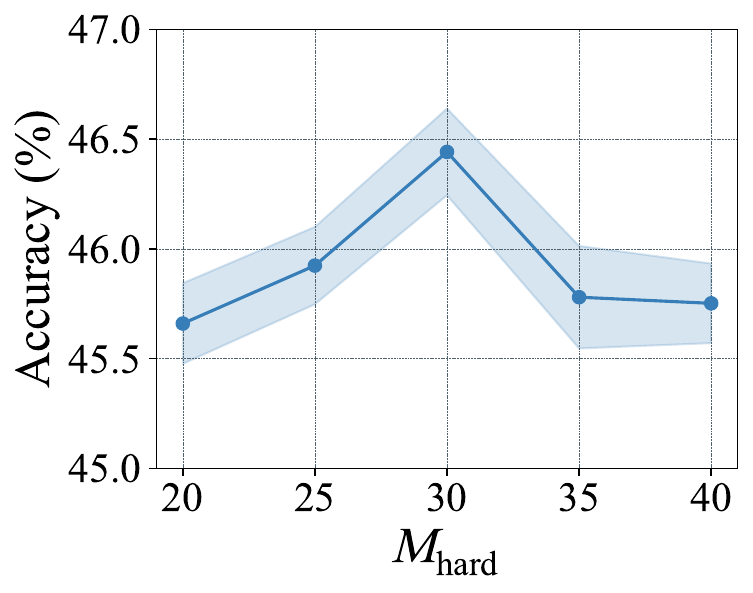}}
	\hfill
     \subfloat[Letter-high]{
     \label{subfig:param2}
     \includegraphics[width = 0.24\textwidth]{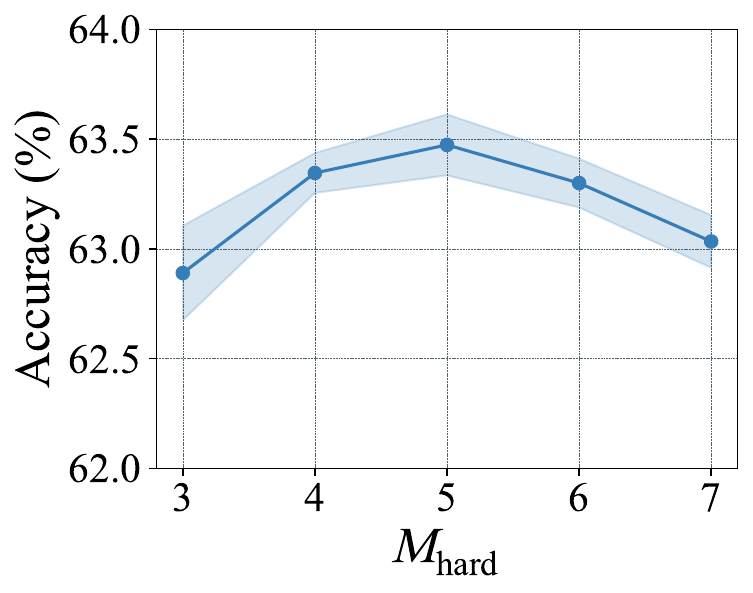}}
	\hfill
     \subfloat[Synthie]{
     \label{subfig:param3}
     \includegraphics[width = 0.24\textwidth]{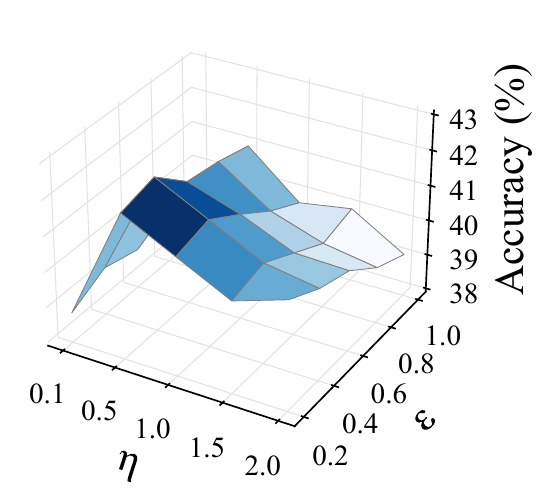}}
	\hfill
	\subfloat[Letter-high]{
    \label{subfig:param4}
     \includegraphics[width = 0.24\textwidth]{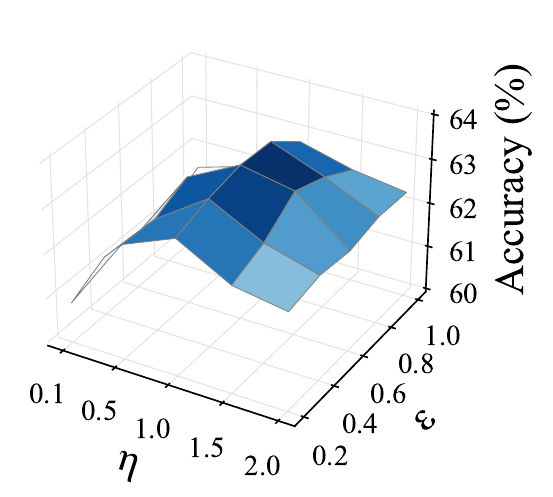}}
\caption{Sensitivity analysis of (a)-(b): hard class number and (c)-(d): loss weight parameters.}
\label{fig:param}
\end{figure*}

\begin{figure}[t]
\centering
\includegraphics[width = 0.35\textwidth]{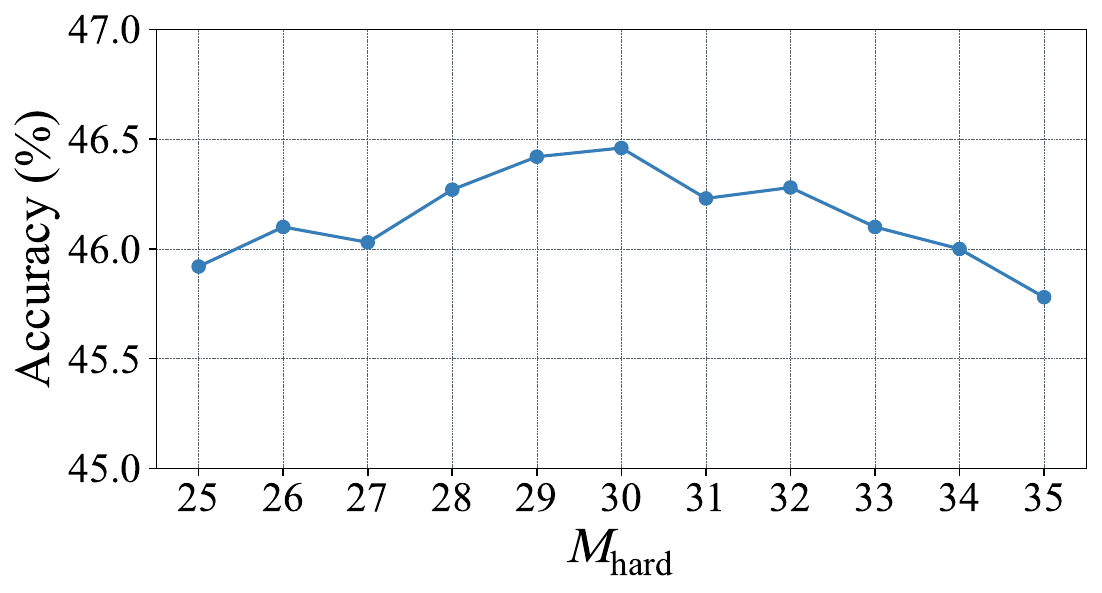}
\caption{Fine-grained sensitivity analysis of hard class number on COIL-DEL (IF=10).}
\label{fig:hcm}
\end{figure}

\subsection{Hyper-parameters Analysis (RQ3)}
In this section, we study the key hyper-parameters in our \method{}, \ie the hard class mining number $M_{\rm hard}$, the balanced contrastive learning weight $\eta$, and the inter-expert distillation weight $\epsilon$. We tune $M_{\rm hard}$ separately by fixing other hyper-parameters, and jointly tune the loss weight hyper-parameters $\eta$ and $\epsilon$. Figure \ref{fig:param} demonstrates how these hyper-parameters affect classification performance.

The impact of $M_{\rm hard}$ on the classification performance of \method{} is illustrated in Figures \ref{subfig:param1} and \ref{subfig:param2}. The experimental results on the COIL-DEL and Letter-high datasets show a similar trend, where the classification accuracy initially improves with an increase in $M_{\rm hard}$. Our model achieves the best performance when the hard class ratio ($M_{\rm hard} / M$) is around 0.3, which corresponds to 30 out of 100 classes for the COIL-DEL dataset and 5 out of 15 classes for the Letter-high dataset. We also conduct more fine-grained adjustments to $M_{\rm hard}$ around 30 on the COIL-DEL dataset in Figure \ref{fig:hcm}. When $M_{\rm hard}$ is set around 30, the differences in performance are minimal, which indicates a relatively consistent and steady trend. 
however, setting $M_{\rm hard}$ to a value that is either too small or too large brings limited gains due to the under- or over-exploration of hard categories.
Further, when dealing with a new dataset, the optimal value of $M_{\rm hard}$ may vary depending on the dataset's specific characteristics. We recommend conducting a small-scale hyper-parameter tuning around $0.3 * M$ to select a suitable $M_{\rm hard}$ for the new dataset.

The loss weight parameters are critical in adjusting the importance of different loss components during training. The impact of $\eta$ and $\epsilon$ is visualized in Figures \ref{subfig:param3} and \ref{subfig:param4}. In our experiments, we vary $\eta \in \{ 0.1, 0.5, 1.0, 1.5, 2.0 \}$ and $\epsilon \in \{ 0.2, 0.4, 0.6, 0.8, 1.0 \}$ on the Synthie dataset (IF=$15$) and Letter-high dataset (IF=$25$). The results indicate that the model is not significantly impacted by the change of $\epsilon$, and the optimal performance is achieved when $\epsilon$ is set to $0.4$ for Synthie and $0.6$ for Letter-high. Moreover, the classification accuracy increases when $\eta$ is raised from $0.1$ to $0.5$, suggesting that balanced contrastive learning significantly contributes to better representation learning.

\begin{table}[t]
\begin{center}
\caption{Effectiveness analysis ($\%$) of disentangled knowledge distillation on the ENZYMES, Letter-high and Letter-low datasets (Target class distillation: TCD; Non-target class distillation: NTCD; Disentangled distillation: DD).} 
\label{table:dkl}
\tabcolsep=1.5pt
{\small
% \resizebox{0.96\columnwidth}{!}{
\begin{tabular}{ccc|cccccc}
\toprule
 \multicolumn{1}{c}{\multirow{2}{*}{TCD}}     &\multicolumn{1}{c}{\multirow{2}{*}{NTCD}}     &\multicolumn{1}{c|}{\multirow{2}{*}{DD}}      &\multicolumn{2}{c}{ENZYMES}  &\multicolumn{2}{c}{Letter-high} &\multicolumn{2}{c}{Letter-low}\\
 \multicolumn{1}{c}{} &\multicolumn{1}{c}{} &\multicolumn{1}{c|}{} &Accuracy &$\Delta$ &Accuracy &$\Delta$ &Accuracy &$\Delta$ \\
\midrule
 & & &37.13 &- &62.85 &- &90.88 &- \\
\Checkmark  &\Checkmark & &37.67 &+0.54 &63.18 &+0.33 &91.20 &+0.32 \\
\Checkmark  & & &37.31 &+0.18 &62.76 &-0.09 &91.11 &+0.23 \\
 &\Checkmark & &37.57 &+0.44 &62.95 &+0.10 &91.31 &+0.43 \\
\Checkmark  &\Checkmark &\Checkmark &38.00 &+0.87 &63.42 &+0.57 &91.67 &+0.79 \\
\bottomrule
\end{tabular}}
\end{center}
\end{table}

\subsection{Effect of Disentangled Knowledge Distillation (RQ4)}

% \subsubsection{Performance gain of each distillation component}
\noindent\textbf{Performance gain of each distillation component.} 
Here we individually study the effects of knowledge distillations on target and non-targets over the ENZYMES, Letter-high, and Letter-low datasets. The accuracy and performance gain are reported in Table \ref{table:dkl}. For each dataset, we present results for five variants of the method, including (1) the vanilla training baseline without inter-expert distillation, (2) classical knowledge distillation using both target and non-target class distillation, (3) target class distillation only, (4) non-target class distillation only, and (5) disentangled distillation. 

\begin{figure*}[t]
\centering
	\subfloat[ENZYMES]{\includegraphics[width = 0.31\textwidth]{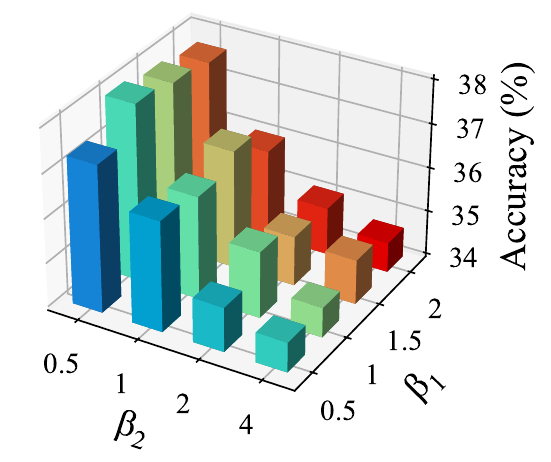}}
	% \hfill
 \quad
	\subfloat[Letter-high]{\includegraphics[width = 0.31\textwidth]{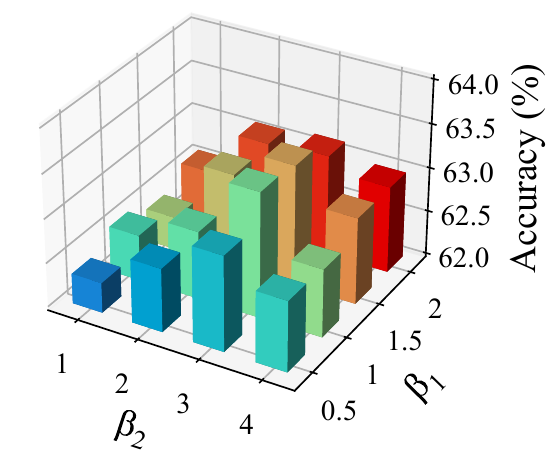}}
 % \hfill
 \quad
	\subfloat[Letter-low]{\includegraphics[width = 0.31\textwidth]{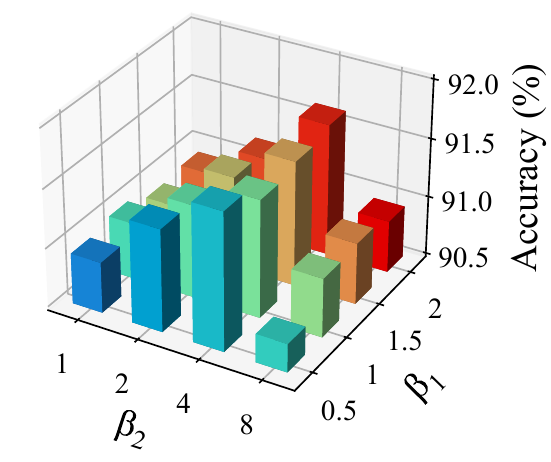}}
\caption{Performance comparison w.r.t. different settings of the distillation weight hyper-parameters.}
\label{fig:dkl}
\end{figure*}

From the table, we can observe that the adoption of classical knowledge distillation leads to an improvement in the overall performance of the model.
% This indicates that inter-expert distillation enables each expert network to learn additional signals from the others, thereby facilitating information sharing among them. 
However, we notice that applying target class distillation solely leads to less improvement or even degrades the overall performance (\eg $0.09\%$ drop on Letter-high dataset). It can also be observed that the performances of using non-target class distillation only are comparable and even better than the classical knowledge distillation (\eg $0.43\%$ accuracy gain on Letter-low dataset). These experimental results illustrate that knowledge related to non-target classes could be more critical than target-class-specific knowledge. Finally, the application of disentangled distillation provides the most significant improvement to our model. This finding suggests that separating different knowledge components in a disentangled manner helps alleviate the issue of information-sharing suppression between target class and non-target classes of different experts.

% \subsubsection{Influence of distillation weight}
\smallskip
\noindent\textbf{Influence of distillation weight.} 
To demonstrate the effectiveness of the distillation weight of target and non-target class knowledge distillation, we carry out an experiment to evaluate the performance under varying distillation weight settings. The experimental results on three datasets are shown in Figure \ref{fig:dkl}. We vary the target class distillation weight $\beta_1$ across $\{ 0.5, 1.0, 1.5, 2.0 \}$ for all three datasets. Additionally, we tune the non-target class distillation weight $\beta_2$ within the range of $\{ 0.5, 1.0, 2.0, 4.0 \}$ for ENZYMES, $\{ 1.0, 2.0, 3.0, 4.0 \}$ for Letter-high, and $\{ 1.0, 2.0, 4.0, 8.0 \}$ for Letter-low.

Overall, the classification performance remains stable when increasing $\beta_{1}$ from $0.5$ to $2.0$. Additionally, we observed that the highest improvement from target class knowledge distillation is achieved when $\beta_{1}$ is set to approximately $1.0$. These results suggest that the model is relatively insensitive to the setting of different $\beta_{1}$ values. However, different settings of $\beta_{2}$ can have a significant impact on the classification performance, and the optimal value of $\beta_{2}$ varies among the different datasets. Specifically, for the ENZYMES dataset, the best performance is achieved when $\beta_{2}$ is set to $0.5$, and increasing the contribution of non-target knowledge distillation results in a drop in accuracy. For the Letter-high and Letter-low datasets, we observed an initial increase in accuracy as $\beta_{2}$ becomes larger, but an excessively large value of $\beta_{2}$ can also lead to performance degradation.
These experimental results demonstrate that the optimal selection of the non-target distillation weight is closely related to the confidence of the experts. When the experts are confident, \ie the predicted probability of the target class is much higher than all non-target classes, the non-target knowledge should be valued more. 
% Thus, the gap between the predicted probability of the target class and the maximum probability among non-target classes can guide the tuning of $\beta_{2}$. 
Specifically, the experts are less confident in the ENZYMES dataset, and the value of $\beta_{2}$ should be set lower than those in the other two datasets. Moreover, a larger value of $\beta_{2}$ could increase the gradient contributed by non-target classes, potentially leading to a degradation of the model's accuracy.
% This is particularly relevant in the case of the Letter-high and Letter-low datasets, where setting an excessively large value of $\beta_{2}$ can have a negative impact on the correctness of prediction.

\section{Conclusion}\label{sec::conclusion}

In this paper, we study long-tailed graph-level classification and propose a novel method termed Collaborative Multi-expert Learning (\method{}) to handle the imbalanced setting of the distribution in graph datasets. \method{} incorporates the balanced contrastive learning for representation learning, accompanied by the classifier learning of each expert to alleviate the influence of the sample sizes in different classes and enhance the hard class mining. To combine the advantages of multiple experts, we design a mechanism to fuse their diversities in a multi-expert framework, thus enhancing the cooperation. Furthermore, we develop a disentangled knowledge distillation to encourage the knowledge transfer and mutual supervision among multiple experts. Extensive experiments demonstrate that our proposed \method{} consistently outperforms the competitive baseline methods on various benchmark graph datasets. 

In our future work, there are several aspects of our proposed model that deserve further investigation: (i) the distribution of the test set is unknown in real-world scenarios, and better mechanisms need to be designed to overcome the unknown distribution rather than the balanced distribution; (ii) extending our framework to more challenging settings such as noisy labels or noisy graphs; (iii) exploring more fundamental theoretical research related to generalization beyond the training distribution such as out-of-distribution problem.
 % (i) exploring better ways to entangle representation learning and classifier learning, so that samples from the tail classes can be better distinguished;

\section*{Acknowledgments}
The authors are grateful to the anonymous reviewers for critically reading the manuscript and for giving important suggestions to improve their paper.

%{\appendix[Proof of the Zonklar Equations]
%Use $\backslash${\tt{appendix}} if you have a single appendix:
%Do not use $\backslash${\tt{section}} anymore after $\backslash${\tt{appendix}}, only $\backslash${\tt{section*}}.
%If you have multiple appendixes use $\backslash${\tt{appendices}} then use $\backslash${\tt{section}} to start each appendix.
%You must declare a $\backslash${\tt{section}} before using any $\backslash${\tt{subsection}} or using $\backslash${\tt{label}} ($\backslash${\tt{appendices}} by itself
% starts a section numbered zero.)}


\bibliographystyle{IEEEtran}
\bibliography{ref.bib}

\begin{IEEEbiography}
[{\includegraphics[width=1in,height=1.25in]{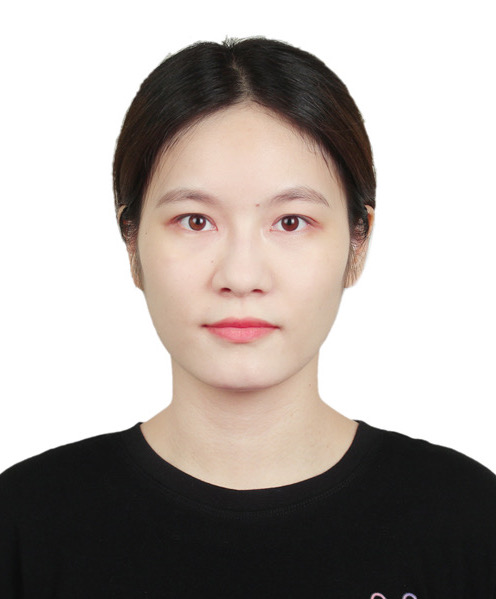}}]
{Siyu Yi} is currently a Ph.D. candidate in statistics from Nankai University, Tianjin, China. She received the B.S. and M.S. degrees in Mathematics from Sichuan University, Sichuan, China, in 2017 and 2020, respectively. Her research interests focus on graph representation learning, design of experiments, statistical sampling, and subsampling in big data. 
\end{IEEEbiography}

\begin{IEEEbiography}
[{\includegraphics[width=1in,height=1.25in]{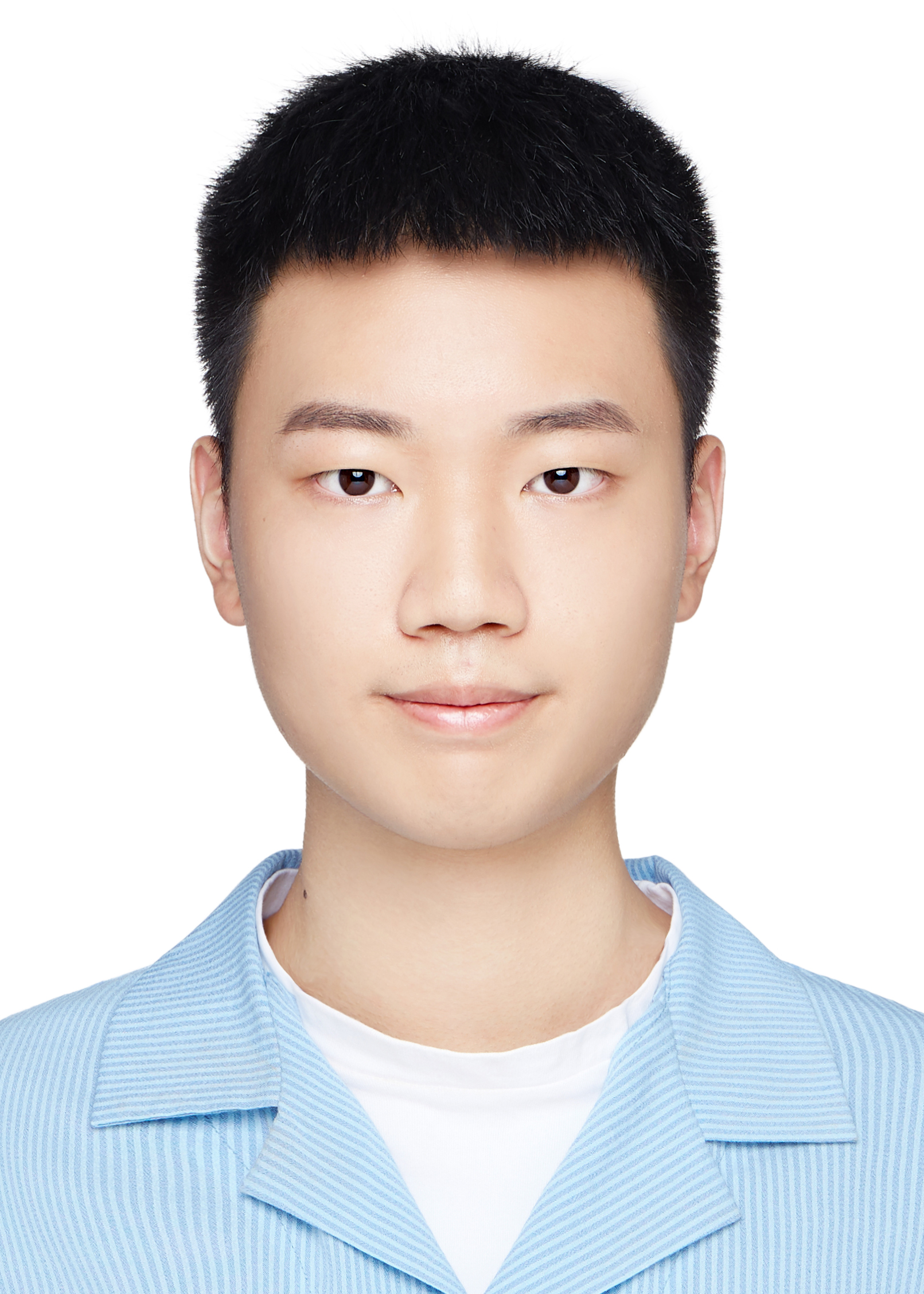}}]
{Zhengyang Mao} is currently a graduate student at the School of Computer Science, Peking University. His research interests lie primarily in the area of machine learning with graph, including graph representation learning, data-imbalanced learning, and semi-supervised learning.
% is an undergraduate student in School of EECS, Peking University, Beijing, China. His research interests include graph representation learning and recommender sysmtes.
\end{IEEEbiography}

\begin{IEEEbiography}
[{\includegraphics[width=1in,height=1.25in]{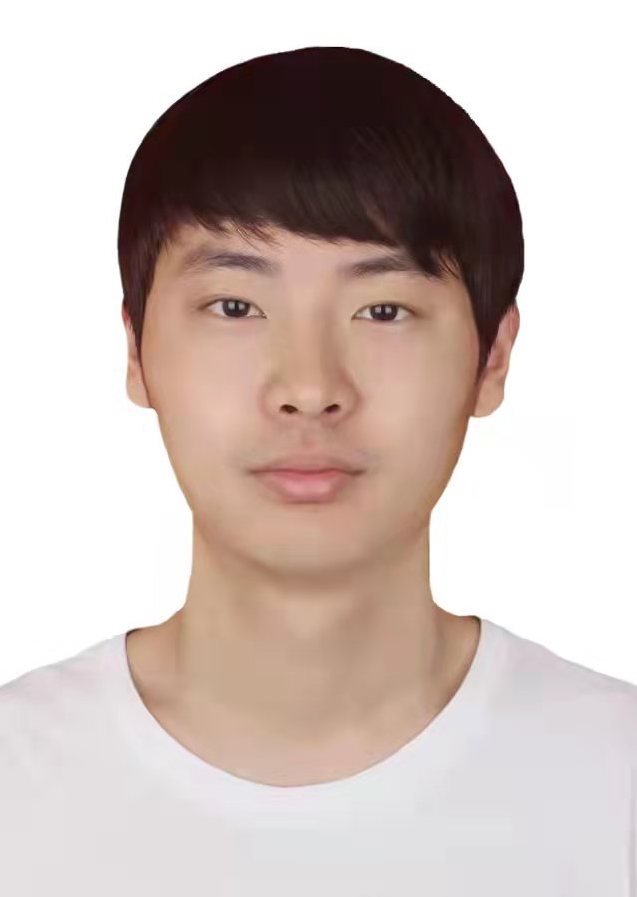}}]
{Wei Ju} is currently a postdoc research fellow in Computer Science at Peking University. Prior to that, he received his Ph.D. degree in Computer Science from Peking University, Beijing, China, in 2022. He received the B.S. degree in Mathematics from Sichuan University, Sichuan, China, in 2017. His current research interests lie primarily in the area of machine learning on graphs including graph representation learning and graph neural networks, and interdisciplinary applications such as knowledge graphs, drug discovery and recommender systems. He has published more than 20 papers in top-tier venues and has won the best paper finalist in IEEE ICDM 2022.
\end{IEEEbiography}

\begin{IEEEbiography}
% ,clip,keepaspectratio
[{\includegraphics[width=1in,height=1.25in]{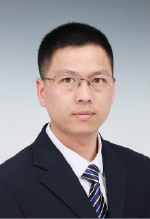}}] {Yongdao Zhou} received his B.S. degree in Mathematics, M.S. and Ph.D. degree
in Statistics from Sichuan University, China, in 2002, 2005, and 2008, respectively.  After graduation, he joined Sichuan University and was a professor after 2015. In 2017, he then joined Nankai University, where he is presently a professor in Statistics. His research agenda focuses on design of experiments and big data analysis. He published more than 60 papers and 5 monographs. His research publications have won best paper awards in WCE 2009 and Sci Sin Math in 2023.
\end{IEEEbiography}

\begin{IEEEbiography}
[{\includegraphics[width=1in,height=1.25in]{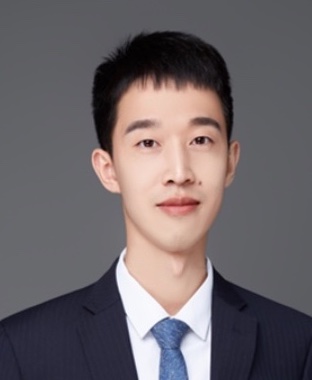}}]
{Luchen Liu} is currently a post-doctoral research fellow in Computer Science at Peking University. He received the Ph.D. degree in Computer Science from Peking University in 2020. His current research interests lie primarily in the area of deep learning for temporal graph data and interdisciplinary applications such as intelligent healthcare and quantitative investment.
\end{IEEEbiography}

\begin{IEEEbiography}
[{\includegraphics[width=1in,height=1.25in]{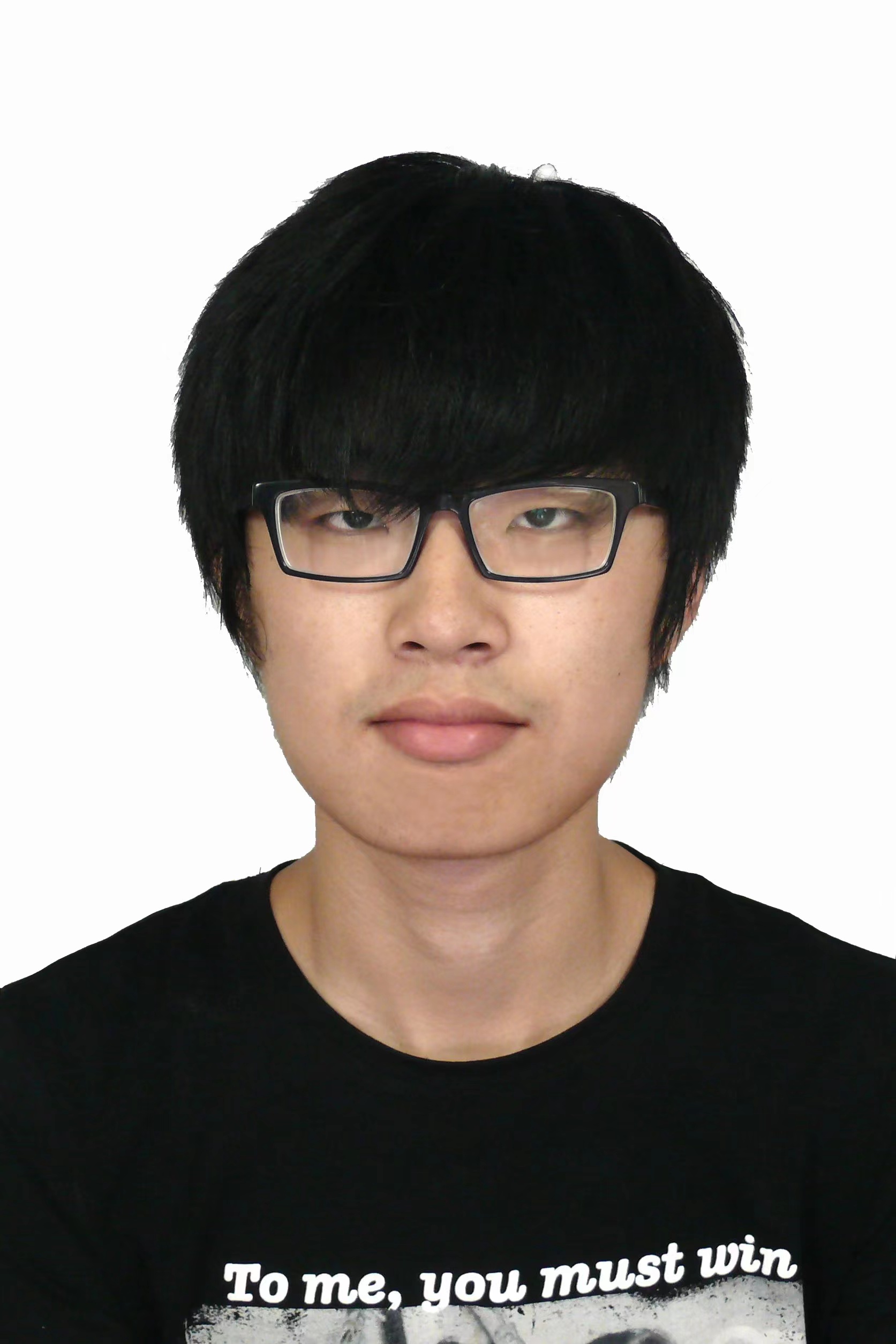}}]
{Xiao Luo} is a postdoctoral researcher in Department of Computer Science, University of California, Los Angeles, USA. Prior to that, he received the Ph.D. degree in School of Mathematical Sciences from Peking University, Beijing, China and the B.S. degree in Mathematics from Nanjing University, Nanjing, China, in 2017. 
His research interests includes machine learning on graphs, image retrieval, statistical models and bioinformatics. 
\end{IEEEbiography}

\begin{IEEEbiography}
[{\includegraphics[width=1in,height=1.25in]{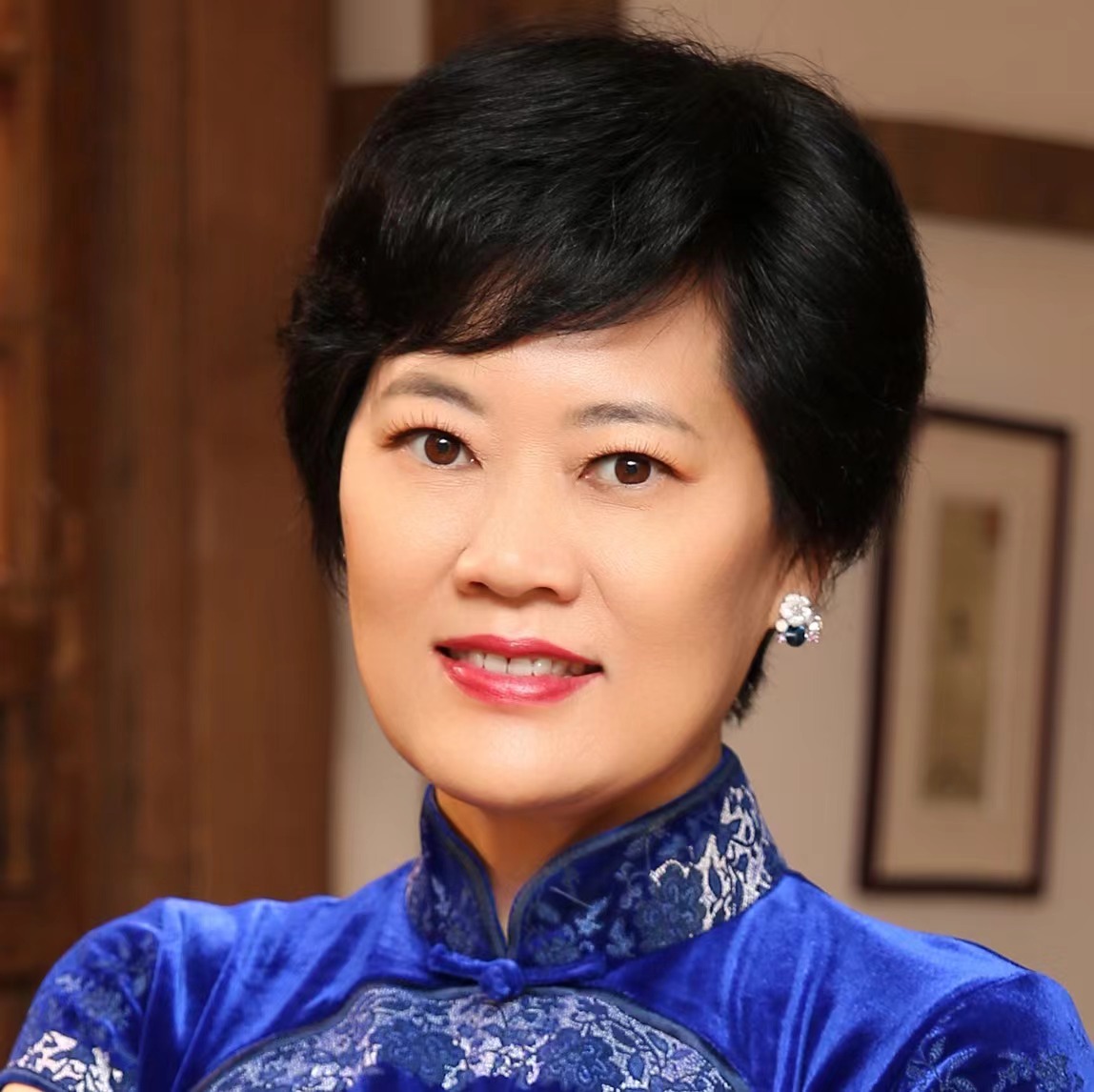}}]
{Ming Zhang} received her B.S., M.S. and Ph.D. degrees in Computer Science from Peking University respectively. She is a full professor at the School of Computer Science, Peking University. Prof. Zhang is a member of Advisory Committee of Ministry of Education in China and the Chair of ACM SIGCSE China. She is one of the fifteen members of ACM/IEEE CC2020 Steering Committee. She has published more than 200 research papers on Text Mining and Machine Learning in the top journals and conferences. She won the best paper of ICML 2014 and best paper nominee of WWW 2016. Prof. Zhang is the leading author of several textbooks on Data Structures and Algorithms in Chinese, and the corresponding course is awarded as the National Elaborate Course, National Boutique Resource Sharing Course, National Fine-designed Online Course, National First-Class Undergraduate Course by MOE China.
\end{IEEEbiography}

\vfill

\end{document}